\pdfoutput=1
\documentclass[11pt]{article}

\usepackage[final]{acl}

\usepackage{times}
\usepackage{latexsym}
\usepackage[ruled,vlined]{algorithm2e}
\usepackage[T1]{fontenc}
\usepackage{amsmath} % for \boldsymbol
\usepackage{bm}      % for \bm
\usepackage[utf8]{inputenc}

\usepackage{microtype}

\usepackage{inconsolata}

\usepackage{graphicx}

\usepackage{pifont}
\usepackage{amsmath}
\usepackage{graphicx}
\usepackage{booktabs}
\usepackage{tikz}
\usepackage{pgfplots} % 画折线图
\usepackage{scalefnt} %用于设置字体
\usepackage{caption}
\usepackage{subcaption}
\usepackage{enumitem}
\usepackage{todonotes}
\usepackage{colortbl}
\usepackage{makecell}
\usepackage{multirow}
\usepackage{xcolor}         % colors
\usepackage{spverbatim}
\usepackage[many]{tcolorbox}
\usepackage{lipsum} % 用于生成示例文字
\usepackage{fancybox}
\usepackage{verbatim}
\usepackage{fancyvrb}

\newcounter{stepcounter}
\newtcolorbox[use counter=stepcounter]{stepbox}[1][]{
    breakable,
    enhanced,
    title=\textbf{Node \thestepcounter},
    colframe=blue!50!black,
    colback=blue!5!white,
    colbacktitle=blue!20!white,
    coltitle=black,
    fonttitle=\bfseries,
    attach boxed title to top left={xshift=5mm,yshift=-2mm},
    boxed title style={colframe=blue!50!black},
    sharp corners,
    before skip=15pt,
    #1
}
\title{FastMCTS: A Simple Sampling Strategy for Data Synthesis}

\author{
 \textbf{Peiji Li\textsuperscript{1,2}\thanks{Equal contribution}},
 \textbf{Kai Lv\textsuperscript{1,2}$^\ast$},
 \textbf{Yunfan Shao\textsuperscript{1,2}},
 \textbf{Yichuan Ma\textsuperscript{1,2}},
\\
 \textbf{Linyang Li\textsuperscript{2}},
 \textbf{Xiaoqing Zheng\textsuperscript{1}},
 \textbf{Xipeng Qiu\textsuperscript{1}},
 \textbf{Qipeng Guo\textsuperscript{2}\thanks{Corresponding Author}}
\\
 \textsuperscript{1}Fudan University
\\
 \textsuperscript{2}Shanghai AI Laboratory
\\
 \small{
  \href{pjli24@m.fudan.edu.cn}{pjli24@m.fudan.edu.cn},
  \href{lilinyang@pjlab.org.cn}{\{lilinyang, guoqipeng\}@pjlab.org.cn}
 }
}
\begin{document}
\maketitle
\begin{abstract}

Synthetic high-quality multi-step reasoning data can significantly enhance the performance of large language models on various tasks. However, most existing methods rely on rejection sampling, which generates trajectories independently and suffers from inefficiency and imbalanced sampling across problems of varying difficulty. In this work, we introduce FastMCTS, an innovative data synthesis strategy inspired by Monte Carlo Tree Search. FastMCTS provides a more efficient sampling method for multi-step reasoning data, offering step-level evaluation signals and promoting balanced sampling across problems of different difficulty levels. Experiments on both English and Chinese reasoning datasets demonstrate that FastMCTS generates over 30\% more correct reasoning paths compared to rejection sampling as the number of generated tokens scales up. Furthermore, under comparable synthetic data budgets, models trained on FastMCTS-generated data outperform those trained on rejection sampling data by 3.9\% across multiple benchmarks. As a lightweight sampling strategy,  FastMCTS offers a practical and efficient alternative for synthesizing high-quality reasoning data. Our code will be publicly released. \footnote{\href{https://github.com/FlyingDutchman26/FastMCTS}{https://github.com/FlyingDutchman26/FastMCTS}}

\end{abstract}

\section{Introduction}

\begin{figure}[t]
    \centering
    \includegraphics[width=\linewidth]{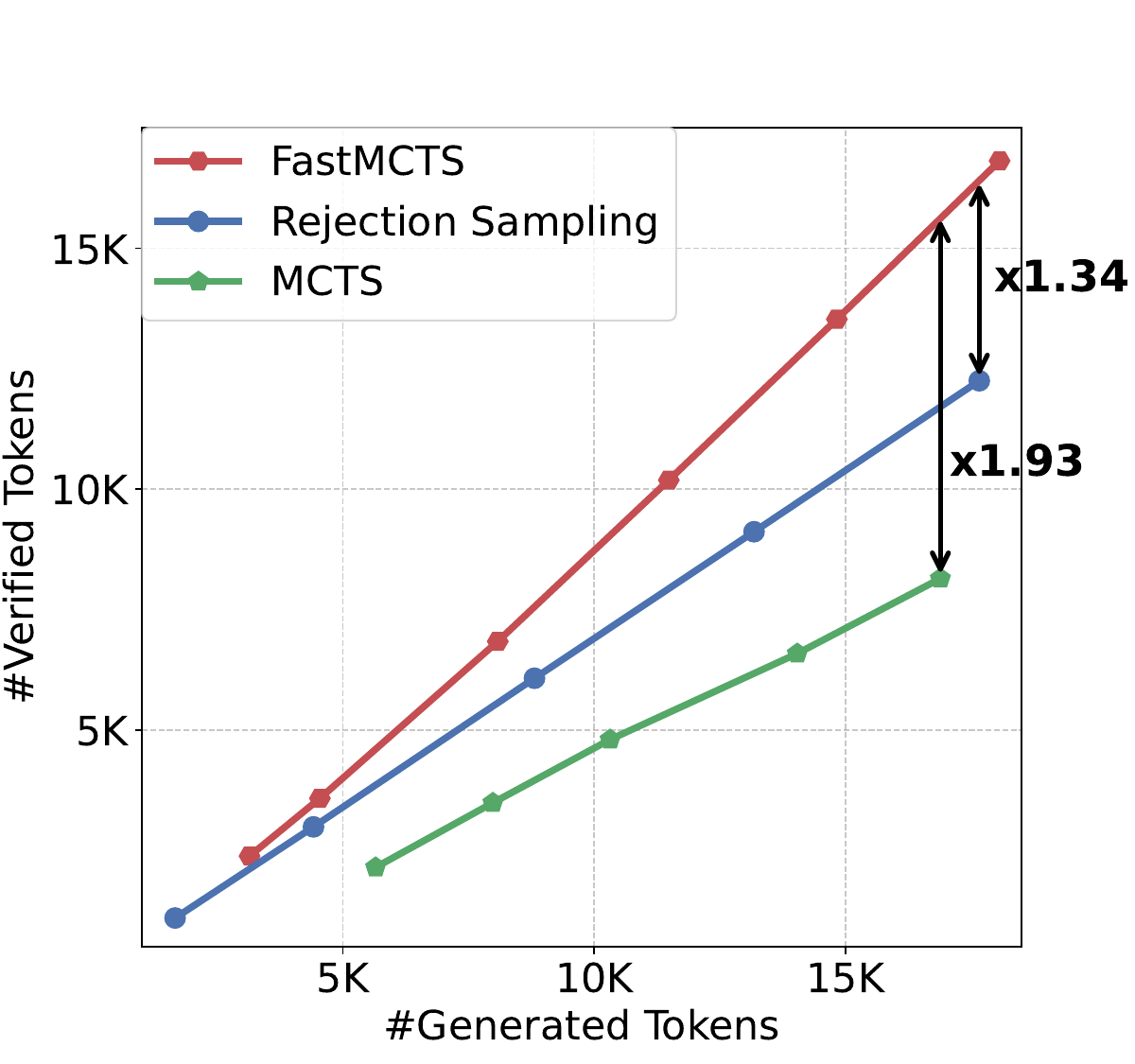}
    \caption{Comparison of generation efficiency of three sampling algorithms. "\#Verified Tokens" represents the total tokens in all verified correct trajectories.}

    \label{fig:intro_fig_cn}
\end{figure}

Large language models (LLMs) have achieved remarkable performance across various domains. 
Reasoning capability plays a crucial role in this success and serves as the foundation for further extending their application scope.
For complex problems, LLMs typically require multi-step reasoning to arrive at final solutions. Synthesizing reasoning trajectories and using them for training has proven to be an effective approach to enhancing their reasoning capabilities.

% 这段讲rejection sampling被用于合成数据时的缺点：采样的独立性以及没有步级别的监督信息。以及引出了MCTS，并指出了MCTS同样不应直接应用于LLM
Currently, rejection sampling~\cite{neal2003slice} is commonly used to synthesize correct trajectories for reasoning tasks. This approach generally involves generating multiple candidate responses through random sampling based on a given problem~\cite{wei2022chain} , and then selecting the correct responses with the corresponding answers as synthetic training data. 
However, this random sampling method handles each attempt independently, constrained by the reasoning capacity of the policy model. As a result, it suffers from inefficiency particularly for long reasoning chains and complex problems, and it fails to provide step-level supervision during the synthesis process.

On the other hand, Monte Carlo Tree Search (MCTS)~\cite{DBLP:conf/cg/Coulom06}, known for its ability to effectively explore state spaces, has been widely adopted in complex tasks such as board games. Some recent studies have also attempted to adapt MCTS for language models. However, the reasoning process of language models differs significantly from those of games like Go or chess. For instance, the state space in language model reasoning is often ill-defined, the computational cost is substantially higher, and the evaluation of reasoning outcomes tends to be more deterministic. As a result, directly applying MCTS to large-scale language generation tasks is less suitable.

% 有待修改

% 这段简单介绍一下我们方法的改进
In this work, we aim to efficiently deploy MCTS for data synthesis. We propose FastMCTS, an MCTS-inspired sample strategy for efficient data synthesis. 
To enhance data synthesis efficiency, we propose a dynamic balance mechanism between exploration and exploitation that adapts to problem complexity.
Specifically, we introduce modifications to the selection phase of MCTS, enabling it to prioritize more valuable nodes rather than being limited to leaf nodes. 
% This search strategy adaptively shifts its focus: when a node has demonstrated successful trajectories in previous simulations, it favors exploration to diversify synthetic reasoning paths; when few or no correct solutions have been discovered through a node, it prioritizes exploitation by exploring proven paths more extensively to identify additional valid solutions.
Furthermore, vanilla MCTS employs a simulation process to evaluate node values. However, conducting complete sampling with LLMs is computationally expensive. To maximize the utility of tokens generated during the autoregressive decoding process of LLMs, we preserve each step of the complete reasoning trajectory generated during simulation as tree nodes, instead of discarding these reasoning steps after simulation. This do not influence the selection of the next most promising node in MCTS but serve as a caching mechanism to prevent redundant generation of reasoning trajectories. Figure \ref{fig:intro_fig_cn} demonstrates the efficiency gains of FastMCTS compared to Rejection Sampling and vanilla MCTS in generating correct trajectory tokens on Chinese high school math data.

% 这段讲了一下实验结果以及我们方法的优越性

Experiments on a wide range of mathematical problems demonstrate the superior data synthesis efficiency of FastMCTS. Compared to vanilla rejection sampling, FastMCTS synthesizes more correct reasoning trajectories, produces more effective tokens, and solves a larger number of problems.
This advantage is particularly pronounced for challenging problems, leading to more balanced synthesis across varying difficulty levels.  Besides, under comparable generation budgets, models trained on FastMCTS-synthesized data outperform those trained on baseline methods across various benchmarks of different complexity.

Further analysis validates the effectiveness of the proposed components and shows that step-level pairwise data constructed through FastMCTS can further boost model performance through methods like step or branch level Direct Preference Optimization. As a lightweight data synthesis strategy, we believe FastMCTS offers a superior alternative to vanilla rejection sampling due to its higher efficiency and ability to provide step-level supervision for multi-step reasoning tasks.

\begin{figure*}[htb]
    \centering
    \includegraphics[width=\linewidth]{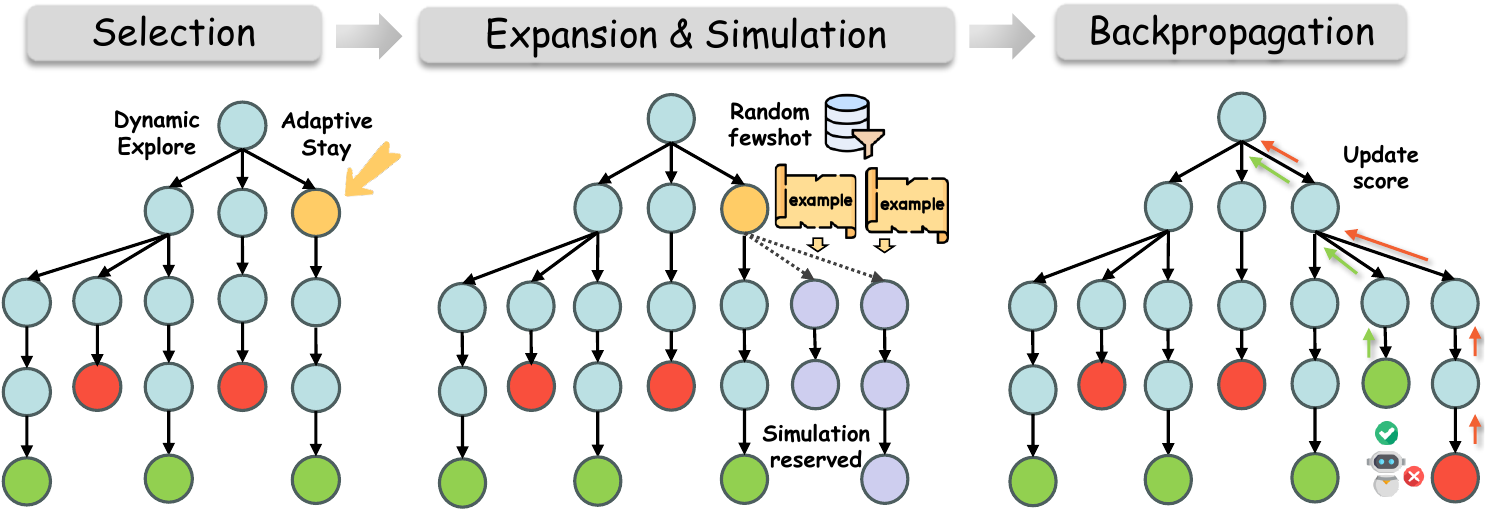}
    \caption{The overview of one iteration of FastMCTS}

    \label{fig:intro}
\end{figure*}

\section{Related Work}

\paragraph{Synthetic Data for Reasoning Tasks}
% \paragraph{Synthetic Data for Math Reasoning }
Synthetic data has become a key resource for improving the reasoning capabilities of large language models. Several studies~\cite{DBLP:conf/iclr/YuJSYLZKLWL24,DBLP:conf/iclr/XuSZG0FTLJ24} focus on generating new problem sets by rephrasing or augmenting existing training data. Other works~\cite{DBLP:journals/corr/abs-2306-02707,DBLP:journals/corr/abs-2403-04706} leverage strong models, such as GPT-4~\cite{Achiam2023GPT4TR}, to distill high-quality reasoning data, enhancing the reasoning capabilities of smaller models; some of these approaches also utilize code executors to further improve performance~\cite{, DBLP:journals/corr/abs-2309-05653,DBLP:conf/iclr/WangRZLLSZSZ024,DBLP:journals/corr/abs-2402-10176}. Additionally, methods like~\cite{DBLP:conf/acl/WangLSXDLCWS24, DBLP:journals/corr/abs-2406-06592,DBLP:conf/emnlp/WangLWLH0S24} focus on synthesizing multi-step reasoning data and provide step-level supervision without the need for human annotation.

\paragraph{Sampling Strategies for Data Synthesis}

Sampling strategies play a crucial role in enhancing the reasoning and generation capabilities of large language models. Many approaches improve reasoning performance by sampling multiple reasoning paths and selecting the most promising ones. For instance, Self-Consistency~\cite{DBLP:conf/iclr/0002WSLCNCZ23} generates diverse reasoning paths and selects the most consistent answer. Other works~\cite{yuan2023scalingrelationshiplearningmathematical,DBLP:journals/corr/abs-2402-10176,DBLP:journals/corr/abs-2407-13690} use strategies like rejection sampling~\cite{neal2003slice} to generates candidate outputs and filters them based on predefined criteria or a reward model. 
% These methods cannot provide effective step-level supervision.

\paragraph{Tree Search in LLM}
Tree-search strategies have been shown to be highly effective in enhancing the reasoning capacity of large language models, as the nodes of the tree can naturally represent reasoning steps in the chain-of-thought (CoT)~\cite{wei2022chain}. Several studies~\cite{yao2024tree,hao2023reasoning,zhang2024accessing,tian2024toward} have employed tree search during inference to guide multi-step reasoning.
% These methods typically prompt large language models to self-evaluate or use reward models to provide process-level supervision, employing depth-first search, breadth-first search, or Monte-Carlo Tree Search strategies. 
In another stream of research~\cite{feng2023alphazero,DBLP:journals/corr/abs-2405-03553,xie2024monte,zhang2024rest,wang2024towards}, Monte-Carlo Tree Search is used to generate tree-structured data for training, constructing preference data pairs or providing process supervision for CoT steps.

However, in synthetic data scenarios of LLMs, using MCTS can incur significant overhead due to simulation costs or rely on a trained process reward model for step supervision, leading to inefficiencies. To address these limitations, we propose FastMCTS, which efficiently synthesizes tree-structured multi-step reasoning data with high efficiency.
% These tree search-based approaches often face low search efficiency.
% These approaches allow for preference optimization or iterative training of the policy model and value/reward model. 

% However, methods based on Rejection Sampling cannot provide effective step-level supervision, while tree search-based approaches often face low search efficiency and heavily rely on reward models. To address these limitations, we propose Fast-MCTS, a reward-model-free tree search strategy that incorporates Monte Carlo Tree Search (MCTS) principles. Our method synthesizes tree-structured multi-step reasoning data more efficiently than vanilla rejection sampling, without requiring additional reward models.

% 这里是不是要列一下我们的改进与contribution

% \paragraph{Reasoning Path Searching and Learning}

% sc 
% tot
% major vote
% best of n 
% rft 
% dpo 
% step-dpo
% grpo
% mcts dpo etc.

% \paragraph{Monte-Carlo Tree Search}

% alpha zero

% mcts in game 

% mcts in LLM 

% rest mcts etc.

\section{Preliminaries}

\paragraph{Rejection Sampling}

Rejection sampling is a widely used synthetic-data method for obtaining high-quality data to enhance the reasoning capabilities of LLMs. 
% This approach is commonly employed to distill knowledge from stronger models, such as GPT-4. 
Given an input question \( q \), the process involves sampling multiple candidate responses \( \{o^{(j)}\}_{j=1}^N \) from a language model. Each response \( o^{(j)} \) is then evaluated based on predefined criteria, typically by comparing its final answer to a ground-truth solution using a rule-based function. Responses that pass this filtering step are considered correct and used to train the language model.

However, vanilla rejection sampling suffers from several limitations. For instance, the sampled data may exhibit imbalanced distributions~\cite{DBLP:journals/corr/abs-2407-13690}. Moreover, due to the rule-based filtering mechanism, reasoning paths with errors in intermediate steps or those incorrectly discarded due to formatting issues are often excluded~\cite{DBLP:conf/iclr/LightmanKBEBLLS24}. Our work addresses these issues effectively by introducing a more robust sampling strategy while achieving higher efficiency.

\paragraph{Monte Carlo Tree Search} 

Monte Carlo Tree Search (MCTS) is a decision-making algorithm widely used in games like Go and complex decision processes~\cite{DBLP:journals/nature/SilverHMGSDSAPL16,DBLP:journals/nature/SilverSSAHGHBLB17}. It builds a search tree through simulations to estimate the value of actions. In the context of language models, MCTS serves as a sampling strategy that can be combined with reward models to assist inference or synthesize multi-step reasoning data, providing step-level supervision for further training.

MCTS iteratively constructs a search tree through four phases: selection, expansion, simulation, and backpropagation~\cite{DBLP:journals/tciaig/BrownePWLCRTPSC12}. When applied to LLM inference, the input question \( q \) is represented as the root node, and each reasoning step in the chain-of-thought (CoT) is represented as a child node. During selection, MCTS uses the Upper Confidence Bound for Trees (UCT) criterion to balance exploration and exploitation:
\begin{equation}
\text{UCT}(i) = \frac{w_i}{n_i} + c \cdot \sqrt{\frac{\ln N_i}{n_i}}  
\label{eq:uct}
\end{equation}
where \( n_i \) is the visit count for node \( i \), \( N_i \) is the visit count for its parent, \( w_i \) is the cumulative value of descendant nodes, and \( c \) is a hyperparameter. 

Unlike board games, each roll-out in language models requires autoregressive inference, making the simulation process computationally expensive~\cite{DBLP:journals/corr/abs-2405-03553}. The results of simulations are often discarded after backpropagation, further reducing sampling efficiency. As a result, directly applying MCTS for data synthesis incurs significant computational overhead.

\section{Method}
% 1. reserve simulation
% 2. stay policy
% 3. dynamic c_puct
% 4. robust: Model Eval + random fewshot

\newcommand{\COMMENTLLAMA}[1]{{\textcolor[HTML]{962C38} {$\triangleright$ {#1}\\}}}
\newcommand{\COMMENTLIGHTGRAY}[1]{\hfill{\textcolor[HTML]{A9A9A9} {$\triangleright$  {#1}\\}}}
\begin{algorithm*}
\caption{Selection phase of FastMCTS}
\label{algo:select}
\KwIn{Current search tree $T$, difficulty thresholds $l_{high}, l_{low}$, UCT constant $c$ }
\KwOut{Selected node in this iteration}
\COMMENTLLAMA{Recursively select node with Adaptive Stay Policy}

% select from root
current\_node $\gets$ root

selected\_node $\gets$ None
    
\While{\textnormal{selected\_node} is None}{

candidate\_children $\gets$ current\_node.children

\If{\textnormal{number of candidate\_children} $<=$ 1 \texttt{or} \COMMENTLIGHTGRAY{\textnormal{Adaptive Stay Policy}}
\quad \textnormal{all candidate\_children are leaf nodes} \texttt{or} \\
\quad current\_node.visit\_count > 1 \texttt{and} current\_node.score $\in (0, l_{\text{low}}] \cup [l_{\text{high}}, 1)$ 
}{
selected\_node $\gets$ current\_node

break
}

\eIf{current\_node.visit\_count > 1}{$c_{current}$ $\gets$ $c \cdot current\_node.score$ \COMMENTLIGHTGRAY{Dynamic Exploration}} 
{$c_{current}$ $\gets$ $c$}

candidate\_node $\gets$ $\arg\max_{node \in candidate\_children} UCT(node,c_{current})$ 

\If{candidate\_node.visit\_count > 1 \texttt{and} candidate\_node.score <= $l_{low}$}{
selected\_node $\gets$ candidate\_node
}

current\_node $\gets$ candidate\_node

}
\end{algorithm*}

\label{method:fastmcts}
% 这段可能去掉或者放在后面，更像是problem formulation
% In our framework for synthetic data generation, we consider an input question as \( q \), and the language model is treated as a policy, denoted as \( \pi_\theta \). 
% Concretely, consider a complete solution consisting of \( T \) reasoning steps. At a given time step \( t \), we represent the partial solution as the state \( s_t \), and the subsequent reasoning step to be taken as the action \( a_{t+1} \). The policy model, implemented by a large language model (LLM), generates the action \( a_t \) based on the current state \( s_t \) and the input question \( q \), i.e.,  
% $
% \pi_\theta(a_{t+1} | s_t, q) = \text{LLM}(a_{t+1} | s_t, q).  
% $  
% The transition from one state to the next is deterministically achieved through the concatenation operation:  
% $
% s_{t+1} = \text{Cat}(s_t, a_t),  
% $  
% where \( s_t = (a_t, a_{t-1}, \dots, a_1) \) represents the sequence of reasoning steps up to time \( t \).
% In our settings, we split the reasoning trajectories to steps by "Step 1", "Step 2" etc.
In our framework for synthetic data generation, for an input question $q$ and its solution with $ T $ reasoning steps, the partial solution at time step $ t $ is represented as state $ s_t $, and the next reasoning step as action $ a_{t+1} $. The language model is treated as a policy model \( \pi_\theta \) and generates actions based on the current state and input question:
\begin{equation}
\pi_\theta(a_{t+1} | s_t) = \text{LLM}(a_{t+1} | s_t)
\end{equation}
The transition to the next state is achieved by concatenating current state and next ction:
\begin{equation}
s_{t+1} = \text{Cat}(s_t, a_{t+1})
\end{equation}
where $ s_t = (a_t, a_{t-1}, \dots, a_1,q) $ represents the sequence of reasoning steps up to time $ t $. We segment the reasoning trajectories into individual steps based on strings such as "Step 1", "Step 2", etc., with each step corresponding to a node in the tree structure. The details of how reasoning steps are separated are provided in Appendix \ref{apd:node}. 

Our proposed method, FastMCTS, introduces several key improvements to vanilla MCTS algorithm, tailored for efficient and robust data synthesis in language models. 
In the following, we describe our algorithm in detail.

\subsection{Selection with Adaptive Stay Policy}

\label{sec:adaptive}

In the selection phase, Fast-MCTS recursively selects child nodes using the Upper Confidence Bound for Trees (UCT) criterion, as vanilla MCTS does, as defined in Equation \ref{eq:uct}. However, to improve efficiency and diversity, we introduce an Adaptive Stay Policy that dynamically adjusts the selection process based on the node's exploration status and estimated value. 

% 我们从合成数据效率的角度出发，尽可能的搜索正确且多样的推理路径。
% 在adaptive Stay policy中，与vanilla MCTS不同，select不再需要选择到叶子节点。对于作对期望较低的state会及时放弃。对于模型作对概率极高或极低的状态，也会stay而不是继续select，而是优先保证多样性，或试图至少探索到一条正确的推理路径。

% In Adaptive Stay policy, selection does not necessarily proceed to leaf nodes as in vanilla MCTS. For states where the model's probability of being correct is either very high or very low, the policy opts to "stay" rather than continuing selection. This approach prioritizes diversity for easy problem and ensures that at least one correct reasoning path is explored for hard problem.

In Adaptive Stay policy, selection does not necessarily proceed to leaf nodes as in vanilla MCTS. For states where the likelihood of being correct is either very high or very low, our method opts to "stay" rather than continuing selection. This approach prioritizes diversity for easier problems and attempts to explore at least one correct reasoning path for more challenging problems.

\begin{figure*}[t]
    \centering
    \begin{subfigure}[b]{1\linewidth}
        \includegraphics[width=\linewidth]{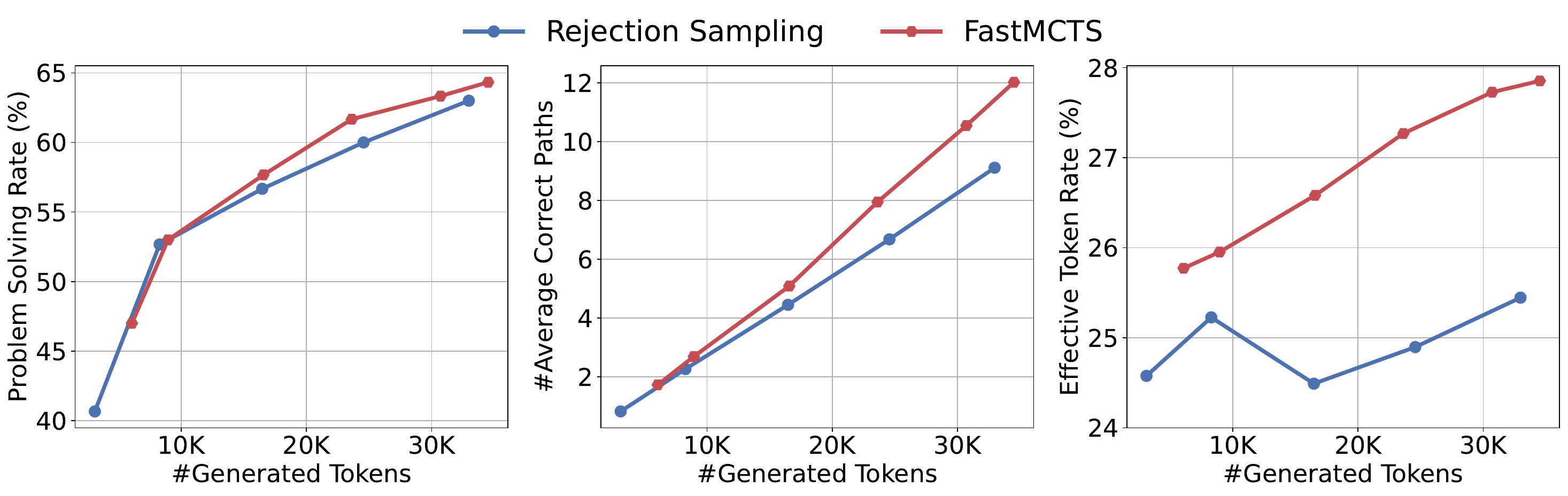}
        \caption{Sampling Efficiency on AIME}
        \label{fig:efficiency_aime}
    \end{subfigure}
    
    % \vspace{1em} % 添加一些垂直间距
    
    \begin{subfigure}[b]{1\linewidth}
        \includegraphics[width=\linewidth]{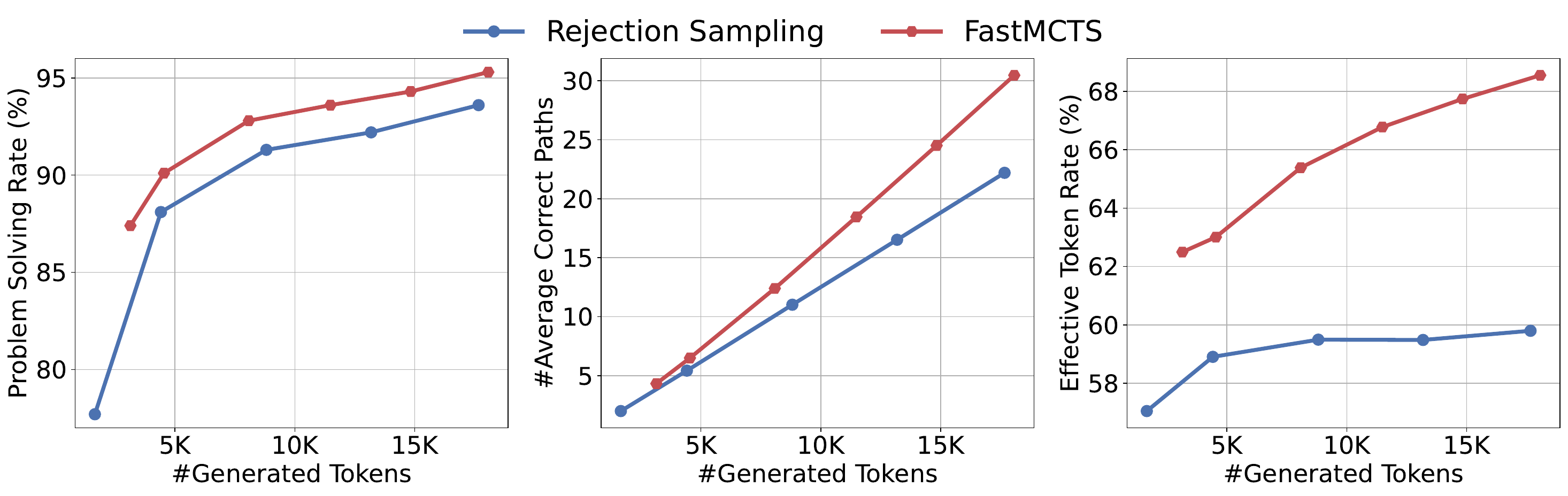}
        \caption{Sampling Efficiency on CN High School Math}
        \label{fig:efficiency_high}
    \end{subfigure}
    
    \caption{Comparison of sampling efficiency for FastMCTS and Rejection Sampling.}
    \label{fig:efficiency}
\end{figure*}

\subsection{Dynamic Exploration}

To enhance the search strategy, we dynamically adjust the parameter $ c $ in UCT based on node scores. The score of one tree node is defined as the estimated value of taking an action (step), calculated by Monte Carlo Evaluation:
\begin{equation}
   node.score = \frac{node.win\_count}{node.visit\_count} 
\end{equation}
Then we adjuct $c$ by multiplying it with the node's score if the node has been visited more than once.
This approach encourages exploration in promising states and prioritizes exploitation in less promising ones, aligning with the goal of data synthesis. The entire selection phase of the FastMCTS algorithm is demonstrated in Algorithm \ref{algo:select}.

% \paragraph{Reserve Simulation}
% To streamline the process and enhance efficiency, we consolidate the expansion and simulation phases into a single phase. From the selected node, we perform \( \text{expand\_degree} \) 次 roll-outs using the LLM policy \( \pi_\theta \), generating \( \text{expand\_degree} \) 个 new reasoning paths. Unlike vanilla MCTS, which discards simulation results, we preserve all newly generated paths as valid nodes and add to our search tree. This ensures that correct reasoning paths are not wasted, significantly improving efficiency. 这个改进能够很好的与之前的Adaptive Stay Policy结合, 因为 保留了Simulation 的推理路径，意味着一个节点即使未被select expand，也会有一条对应的子孙孩子节点分支对应着一条推理路径。因此select过程是不需要深入到叶子节点。

\subsection{Reserve Simulation}
% 参考一些motivation部分

% Unlike board games such as Go or chess, where the outcome of a random simulation does not necessarily reflect the quality of a specific node, LLM reasoning exhibits a strong correlation between the final answer and the correctness of the entire reasoning path. In most cases, if the final answer is correct, the entire reasoning path is likely to be correct. Therefore, simulation results in LLMs are valuable and should be preserved, rather than discarded as in traditional MCTS. This fundamental difference makes the direct application of vanilla MCTS strategies unsuitable for LLM reasoning tasks.

% Unlike board games such as Go or chess, where the outcome of a random simulation does not necessarily reflect the quality of a specific node, LLM reasoning exhibits a strong correlation between the final answer and the correctness of the entire reasoning path. Therefore, simulation results in LLMs are valuable and should be preserved, rather than discarded as in traditional MCTS.

% 受此启发，we consolidate expansion and simulation into a single phase. Unlike vanilla MCTS, which discards simulation results, we preserve all newly generated paths as valid nodes and add them to our search tree. This ensures that all correct reasoning paths are not wasted, significantly improving efficiency. This improvement also integrates well Adaptive Stay Policy, as all trajectories are stored after the selection process, then there is no  need to dive deeply into leaf nodes in search process.

Unlike board games like Go or chess, where the outcome of one random simulation does not necessarily reflect the quality of a specific state, LLM reasoning shows a strong correlation between the final answer and the correctness of the entire reasoning path. Therefore, simulation results in LLMs are valuable and should be preserved.

Inspired by this, we consolidate expansion and simulation into a single phase. Unlike vanilla MCTS, which discards simulation results, we preserve all newly generated paths as valid nodes and add them to our search tree. This significantly enhances sampling efficiency and integrates well with Adaptive Stay Policy. Since all trajectories are stored after selection, there is no need to delve deeply into leaf nodes during the search process.

\subsection{Robustness Enhancements}
\label{sec:robust}
% To address the challenges of answer format variability and logical errors in reasoning paths, we introduce a robustness enhancement mechanism. Instead of relying solely on rule-based answer matching, we use LLM to evaluate the correctness of reasoning paths against the ground-truth answer. Additionally, we require  the correctness of intermediate steps within each reasoning path, ensuring that both the final answer and the intermediate logic are accurate.

% Furthermore, To increase the diversity of generated reasoning paths, we prepend a random set of few-shot examples to the input question during each simulation. Specifically, ext learning (ICL) approach encourages the generation of diverse reasoning paths, further enhancing the robustness of the synthesized data.

To address variability in answer formats and logical errors in reasoning paths, we introduce a robustness enhancement mechanism. Instead of relying solely on rule-based answer matching, we use a LLM to evaluate the correctness of reasoning paths against the ground-truth answer. Additionally, we require the LLM to verify the correctness of intermediate steps within each path, aiming to identify logical errors and exclude trajectories that are guessed answers (e.g., multiple-choice questions). Details of our LLM evaluation methods are described in Appendix \ref{apd:judge}.

Furthermore, to increase the diversity of generated reasoning paths, we prepend different random combinations of few-shot examples to each input string during simulation. To ensure a balanced distribution across mathematical disciplines, we constructed diverse exemplar sets for both Chinese and English datasets, covering domains such as trigonometry, analytic geometry, conic sections, derivatives, calculus, number theory, discrete mathematics, and linear algebra, ensuring sufficient diversity in prompt initialization. Each exemplar was standardized to enforce multi-step reasoning with explicit intermediate steps labeled as "Step 1", "Step 2", etc. This in-context learning approach promotes diverse reasoning paths, further enhancing data robustness.

\subsection{Tree Construction and Data Utilization}
The search tree is constructed iteratively, starting from the root node. The complete algorithm is outlined in Appendix \ref{apd:fastmcts}, and Figure \ref{fig:intro} illustrates the flow of one iteration of FastMCTS.

We can construct training data from the tree structure. Specifically, correct reasoning paths are used for Supervised Fine-Tuning (SFT). Additionally, different branches within the tree nodes, based on their values, can be transformed into pair data for step-level and branch-level Direct Preference Optimization~\cite{DBLP:conf/nips/RafailovSMMEF23}.

\section{Experiment}
% 首先是关于效率的实验
% 其次是关于效果的实验 包括dpo数据的利用
% 最后abalation关于多样性和难度平衡

% In this Section, 我们首先将在\ref{sec:efficiency}展示FastMCTS相比于Rejection Sampling在合成数据效率上的优势。在\ref{sec:performance}中，我们展示了FastMCTS在与Rejection Sampling在相同的推理开销下合成的数据集，在训练效果上的提升。在\ref{sec:analysis}中，我们用消融实验证明了我们方法改进之处的比较性，并且对FastMCTS合成数据的效果进行了分析。

% Todo: 要在这里提一下，为了更广泛的验证我们的方法有效性，我们在两种不同分布的数据上做了效率的验证以及训练的验证

% In this section, we first demonstrate the efficiency advantages of FastMCTS over Rejection Sampling in synthesizing data in Section \ref{sec:efficiency}. 

% In Section \ref{sec:performance}, we demonstrate the improvements in training performance when using datasets synthesized by FastMCTS compared to those generated by Rejection Sampling, under the same inference cost (i.e., generating the same number of tokens). 

% Finally, in Section \ref{sec:analysis}, we conduct ablation studies to highlight the comparative improvements of our approach and provide an in-depth analysis of the synthesized data quality produced by FastMCTS.

% \subsection{Setup}
% data: hard questions from NuminaMath. qwen2.5-72b for roll out. 

% for efficiency, we compare in \S\ref{sec:efficiency}

% for performance, we train qwen-2.5-7B. evaluated on 3 types of benchmark: competition level, college level, olympiad level. compare 7B-math(math-specilaized model), 72B-instruct(roll out model), numinamath-7b. in \S\ref{sec:performance}

\subsection{Sampling Efficiency Comparison}

\label{sec:efficiency}
% In this section ,我们展示了FastMCTS相比于Vanilla Rejection Sampling在Sampling Efficiency上的提升。We use Qwen2.5-72B-Instruct(cite) as the policy model 用于synthetic data，数据集方面，我们使用了AIME(2023年前)的数据，以及从互联网中收集的中文高中数学数据(cite 文曲星)。我们比较了两种方法合成正确题目的效率。

% In this section, we demonstrate the improvements in sampling efficiency of FastMCTS compared to Vanilla Rejection Sampling. For the dataset, we utilized problems from the USA Mathematical Olympiad-level competition AIME up to the year 2023~\cite{aime}, along with subsets of Chinese high school mathematics problems collected from internet \cite{2024internlm2wqx}. We compare the efficiency of both methods in generating correct problem instances.

% In our Experiment, we use sglang(cite) as our 推理引擎， we employ sampling generation with 
% temperature = 1 to ensure diversity, 对于UCT分数中的常数c我们设置为默认值1.414。We also use Qwen2.5-72B-Instruct as a LLM judger to verify the solution. 这能够有效避免由于格式问题引起的判别错误，并能排除一些中间过程错误但做对的推理路径(如选择题)。 To scale up sampling compute, for FastMCTS, 我们逐步扩大迭代的轮数(从4轮扩展到18轮以上)。for Rejection Sampling，我们扩大采样次数，从3扩大到32。 

In this section, we demonstrate the improvements in sampling efficiency of FastMCTS compared to Rejection Sampling. For our dataset, we utilized problems from the USA Mathematical Olympiad-level competition AIME up to the year 2023~\cite{aime}, along with Chinese high school mathematics problems collected from the internet, referred to as CN High School Math \cite{2024internlm2wqx}. Specifically, we randomly selected 300 problems from AIME and 1000 problems from CN High School Math for our experiments. We then compared the efficiency of both methods in generating correct problem instances. We use the open-sourced LLM Qwen2.5-72B-Instruct~\cite{DBLP:journals/corr/abs-2412-15115} and temperature is set to 1. Detailed generation settings are provided in Appendix \ref{apd:sampling settings}.
% 补充在这里

% On these two datasets, our experiment result is in \ref{fig:efficiency} 所示. we scale up generated tokens in our sampling, and we compare three metrics for Rejection sampling and our FastMCTS. Problem Solving Rate 指的是对于对于一个query，合成了至少一条正确推理路径的平均概率。Correct Path 指的是对于一个query，平均产生了正确推理路径的数量。Effective Token Rate指的是生成的token中属于正确推理路径的token的比例。

% 可以看到，随着采样次数增多，generated tokens 数量上升，两种sampling 策略产生的正确推理路径数量都在增加。但是在token数量足够多时，FastMCTS能够比Rejection Sampling多采样出超过30\%的正确推理路径。证明了我们方法在合成数据上的效率提升。此外，FastMCTS也比Rejection Sampling有更高的Problem Solving Rate。，这是由于在每次simulation过程前，对于expand的不同分支都会拼接不同的few-shot作为context，使产生的推理路径更具多样性，搜索到正确答案的概率也得到提升。

Our experimental results are shown in Figure \ref{fig:efficiency}. We gradually increased the number of generated tokens during sampling and compared three metrics for Rejection Sampling and FastMCTS.
\textbf{Problem Solving Rate} refers to the average probability of generating at least one correct reasoning trajectories for a query.
\textbf{Average Correct Paths} refers to the average number of correct reasoning trajectories generated for a query.
\textbf{Effective Token Rate} refers to the proportion of generated tokens that belong to correct reasoning trajectories.

% As shown in Figure \ref{fig:efficiency}, when the number of generated tokens scales up, FastMCTS can generate over 30\% more correct reasoning paths compared to Rejection Sampling, as well as more effective tokens. This demonstrates the efficiency improvement of FastMCTS in synthesizing data. Additionally, FastMCTS exhibits a higher Problem Solving Rate than Rejection Sampling. This is due to the fact that different few-shot examples are prepended as context for each expanded branch before each simulation, enhancing the diversity of generated reasoning paths and increasing the probability of finding the correct solution.

As shown in Figure \ref{fig:efficiency}, FastMCTS generates over 30\% more correct reasoning paths compared to Rejection Sampling as the number of generated tokens scales up. Additionally, FastMCTS produces more effective tokens, demonstrating its efficiency in data synthesis. Furthermore, FastMCTS achieves a higher Problem Solving Rate than Rejection Sampling. This is because diverse few-shot examples are prepended as context for each expanded branch before simulation, enhancing the diversity of generated reasoning paths and increasing the likelihood of finding the correct solution.

\begin{table}[t]
\footnotesize
\centering

\begin{tabular}{lcc}
    \toprule
    &  \multicolumn{1}{c}{\textbf{Rejection Sampling}} & \multicolumn{1}{c}{\textbf{FastMCTS}} \\
    \midrule
    \multicolumn{3}{c}{\textit{EN Math Hard}}  \\
    \# Tokens &  27.8K & 26.2K  \\
    \# Trajectories &  3.46  & \textbf{5.88} \\
    \midrule
    \multicolumn{3}{c}{\textit{CN High School Math Hard}}  \\
    \# Tokens &   18.2K  & 17.4K \\
    \# Trajectories &  8.15 & \textbf{13.70} \\
    \bottomrule
\end{tabular}
\caption{Comparison of synthetic data generation costs between Rejection Sampling and FastMCTS under the experimental settings of Section \ref{sec:performance}. The row ``\# Tokens'' indicates the average number of tokens generated per problem during the sampling phase. The row ``\# Trajectories'' indicates the average number of correct reasoning paths acquired per problem.}

\label{tab:sample_set}
\end{table}

\begin{table*}[htb]
\footnotesize
% 好东西 控制行间距
% \renewcommand{\arraystretch}{1.4}

\begin{tabular}{l@{\hskip 5pt}c@{\hskip 0pt}cccc@{\hskip 5pt}c@{\hskip 5pt}cc@{\hskip 5pt}cc}
\toprule
                     & & \multicolumn{1}{c}{\textbf{Base Level}} & \multicolumn{2}{c}{\textbf{High School Level}}                                                                  & \multicolumn{3}{c}{\textbf{Competition Level}} & \multicolumn{2}{c}{\textbf{Olympiad Level}}                         &  \\ 
                     \cmidrule(lr){3-3} \cmidrule(lr){4-5} \cmidrule(lr){6-8} \cmidrule(lr){9-10} 
Method &    \#Data   & GSM8K                     & \begin{tabular}[c]{@{}c@{}}Gaokao\\ Math\end{tabular} & \begin{tabular}[c]{@{}l@{}}SAT\\ Math\end{tabular} & AIME24          & AMC23         & MATH         & \begin{tabular}[c]{@{}c@{}}Olympiad\\ Bench\end{tabular} & OmniMath & \textbf{Avg.}\\ \midrule
Qwen2.5-7B & - & 88.2 & 62.6    &  70.6 &  0 &  47.5 &  66.8 &    26.2 & 35.5 & 49.7 \\ \midrule
\multicolumn{11}{c}{\textit{Training Trajectories per Problem $\le$ 5}}                                                                                                                                                                                              \vspace{1.5pt}               \\  
RS  & 111K  & \underline{89.1} & 62.6 &  70.6 & 6.7 &  52.5 & 72.0 &    27.6 & 38.3 & 52.4 \\
FastMCTS & 132K & 88.9 & \underline{63.6} &  \underline{74.5} & \underline{13.3} &  \underline{57.5} & \underline{73.0} &    \underline{28.1} & \textbf{39.8} & \underline{54.8}     \\ \midrule
\multicolumn{11}{c}{\textit{Training Trajectories per Problem $\le$ 10}}                                                                                                                    \vspace{1.5pt}                                                                                                 \\
RS  & 167K & 89.4 & 62.6 &  72.6 & 6.7 &  50.0 & 70.8 &    26.3 & 37.5 & 52.0 \\
FastMCTS &223K& \textbf{90.0} & \underline{64.0} &  \underline{74.5} & \underline{13.3} &  \underline{57.5} & \underline{72.0} &  \underline{27.3} & \underline{38.7} & \underline{54.7}    \\ \midrule
\multicolumn{11}{c}{\textit{Training Trajectories per Problem $\le$ 16}}                                                            \vspace{1.5pt}                                                                                                                                                         \\
RS &197K  & 87.1 & \textbf{65.1} &  72.6 & 10.0 &  52.5 & 70.0 &  27.1 & 37.2 & 52.7 \\
FastMCTS &288K & 88.9 & 63.8 &  72.6 & \textbf{20.0} & \textbf{60.0} & 74.0 &  27.5 & 38.3 & 55.6  \\ 
+ Branch-DPO & 152K & \underline{89.9} & 65.0 &  \textbf{76.5} & \textbf{20.0} & 57.5 & \textbf{75.4} &  \textbf{29.6} & \underline{39.2} & \textbf{56.6} \\
\bottomrule
\end{tabular}

% \caption{Comparison of training performance on some challenging math benchmarks benchmarks between Rejection Sampling and FastMCTS with comparable synthetic data generation cost.}
\caption{The results of model performance trained on EN Math Hard dataset synthesized by Rejection Sampling and FastMCTS with comparable generation cost. RS refers to synthetic dataset generated though rejection sampling. \textbf{Bold} indicates the best value, and \underline{underlined} indicates the best value within a group.}

\label{tab:train res en}
\end{table*}

\begin{table}[htb]
% \footnotesize
% 好东西 控制行间距
% \renewcommand{\arraystretch}{1.4}

\begin{tabular}{lccc}
\toprule
Method       & \#Data & Gaokao24 & CMATH \\
\midrule
Qwen2.5-7B   & -  &  33.3   &  85.8     \\
\midrule
\multicolumn{4}{c}{\textit{Training Trajectories per Problem $\le$ 5}}   \\
RS           &  158K      &   58.0   &  89.3     \\
FastMCTS     &  198K      &  \underline{59.4}    &  \textbf{90.8}     \\
\midrule
\multicolumn{4}{c}{\textit{Training Trajectories per Problem $\le$ 10}}   \\
RS           &  250K      &  59.4  &   89.3    \\
FastMCTS     &  359K      &  \underline{60.9}   &   \underline{89.5}    \\
\midrule
\multicolumn{4}{c}{\textit{Training Trajectories per Problem $\le$ 16}} \\  
RS           &  305K      &  60.9        & 88.8      \\
FastMCTS     &  502K      &  \textbf{62.3}        &  89.3     \\
+ Branch-DPO &  215K      &  \textbf{62.3}        &  \underline{89.8}     \\
\bottomrule
\end{tabular}

% \caption{Comparison of training performance on some challenging math benchmarks benchmarks between Rejection Sampling and FastMCTS with comparable synthetic data generation cost.}
\caption{The results of model performance trained on CN High School Math Hard dataset synthesized by Rejection Sampling and FastMCTS with comparable generation cost. RS refers to synthetic dataset generated though rejection sampling. \textbf{Bold} indicates the best value, and \underline{underlined} indicates the best value within a group.}

% 是不是要写一下训练参数？？？？
\label{tab:train res cn}
\end{table}

% \paragraph{Difficulty Level}
% 这个放到analysis

\subsection{Training Performance Comparison}
\label{sec:performance}

% 这些部分删一下，放附录。正文说清楚，细节放附录
\subsubsection{Experimental Setup}
In addition to the comparison of sampling efficiency, we also evaluated the training performance on datasets generated using FastMCTS versus those generated using Rejection Sampling, with comparable computational budgets. To facilitate a more general comparison, we conducted experiments on datasets with two different distributions, specifically Chinese and English.

% We synthesized corresponding mathematical reasoning data for both English and Chinese datasets using these two methods.
% 为了使，我们在
% We refer to these two datasets after selection as \textit{EN Math Hard} and \textit{CN High School Math Hard}.

\paragraph{Training Data Generation} For English data, we selected 46,000 problems from a wide range of math data including Numina-Math~\cite{numina_math_datasets}, MetaMath~\cite{yu2023metamath}, and the training set of InternLM-Math~\cite{ying2024internlmmathopenmathlarge}. For Chinese data, we selected 50,000 problems from Chinese high school math problems collected from the Internet\cite{2024internlm2wqx}. We used heuristic strategies and model evaluations to filter out simpler problems, retaining multiple-choice, fill-in-the-blank, and solution-type questions while excluding proof and diagram-drawing problems. More details for our data selection process are provided in Appendix \ref{apd:data}. We refer to these two datasets after selection as \textit{EN Math Hard} and \textit{CN High School Math Hard}. We used Qwen2.5-72B-Instruct as the policy model and other sampling settings are described in Appendix \ref{apd:sampling settings}. To ensure a fair comparison, we controlled the computational costs of both sampling strategies to be comparable. The specific computational costs for both datasets are detailed in Table~\ref{tab:sample_set}. Under this configuration, FastMCTS generates fewer tokens per query while acquiring more correct reasoning trajectories compared to rejection sampling. 

\paragraph{Baselines}
We use Qwen2.5-7B~\cite{DBLP:journals/corr/abs-2412-15115} and compare its performance when trained on data synthesized by FastMCTS and Rejection Sampling. For both methods, synthesized data is constructed into supervised fine-tuning datasets by randomly sampling different maximum limits of correct trajectories. For FastMCTS, we additionally construct preference data from its tree structures, including step-level and branch-level pairs, which are used for a second-phase Branch-DPO training. Detailed data construction and training setups are provided in Appendix \ref{apd:data construct} and \ref{apd:training details}.

% 评测集 中英

\subsubsection{Main Results}

We evaluated our models across a variety of mathematical benchmarks. All models are assessed in a zero-shot setting, employing greedy decoding for evaluation purposes.

For models trained on data synthesized from EN Math Hard, we evaluated on GSM8K~\cite{DBLP:journals/corr/abs-2110-14168} for baseline assessment, Gaokao Bench Math~\cite{DBLP:conf/icml/TangZWW24} and SAT-Math~\cite{DBLP:conf/icml/TangZWW24} for high school-level problems, AIME24~\cite{aime24}, AMC23~\cite{amc}, and MATH-500~\cite{hendrycksmath2021,DBLP:conf/iclr/LightmanKBEBLLS24} for competition-level challenges, and Olympiad Bench~\cite{DBLP:conf/acl/HeLBHTSHHHZLQL024} and OmniMath~\cite{DBLP:journals/corr/abs-2410-07985} for olympiad-level tasks. For models trained on CN High School Math Hard, we evaluated on 69 text-only problems from the 2024 Chinese Gaokao(National Higher Education Entrance Examination) and CMATH~\cite{wei2023cmath} for foundational performance. Our training data are carefully curated to ensure no overlap with these evaluation benchmarks.

The training results are presented in Table~\ref{tab:train res en} and Table~\ref{tab:train res cn}. Key findings include:

1. Under comparable generation budgets, models trained on FastMCTS-sampled data consistently outperform those trained on rejection sampling data.

2. The performance of models trained on FastMCTS-generated data improves as the number of reasoning trajectories per problem increases, while models trained on rejection sampling data show limited and inconsistent improvement.

3. FastMCTS-generated data can be effectively reused for Branch-DPO training, further enhancing reasoning performance.

These results demonstrate that FastMCTS-synthesized data is more effective than rejection sampling, even with a comparable or lesser generation budget. For FastMCTS, model performance improves with an increase in the number of trajectories used for training, and additional gains can be achieved through DPO by utilizing step-level scores from tree-structured data.

To further validate the effectiveness and robustness of our methods, we also conducted experiments on models of different series with different parameter sizes. Results could be found in Appendix \ref{apd:different models}.

\subsection{Analysis}
\label{sec:analysis}

\begin{figure*}[htb]
    \centering
    \begin{subfigure}[b]{0.48\linewidth}
        \includegraphics[width=\linewidth]{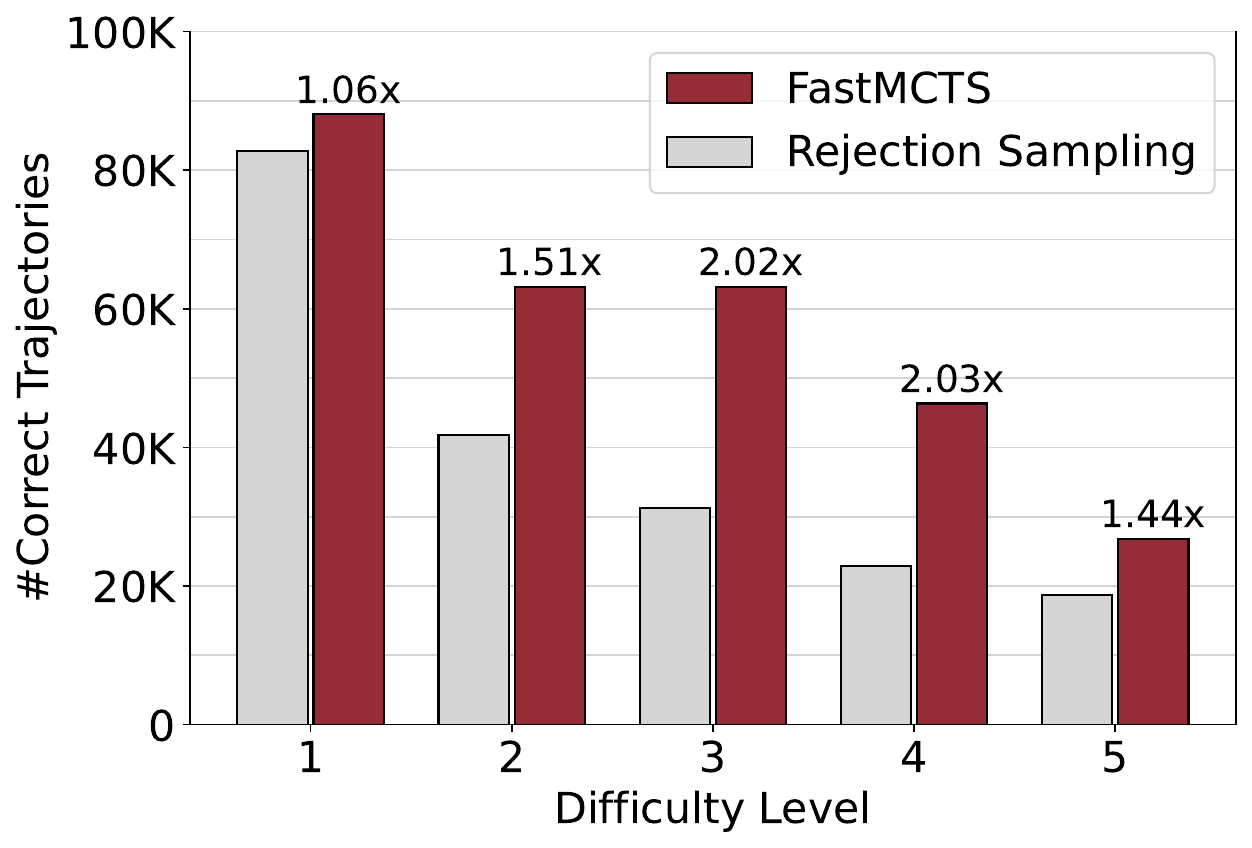}
        \caption{Sampling Balance on EN Math Hard}
        \label{fig:balance_en}
    \end{subfigure}
     % \vspace{1em} 
    \hfill
    \begin{subfigure}[b]{0.48\linewidth}
        \includegraphics[width=\linewidth]{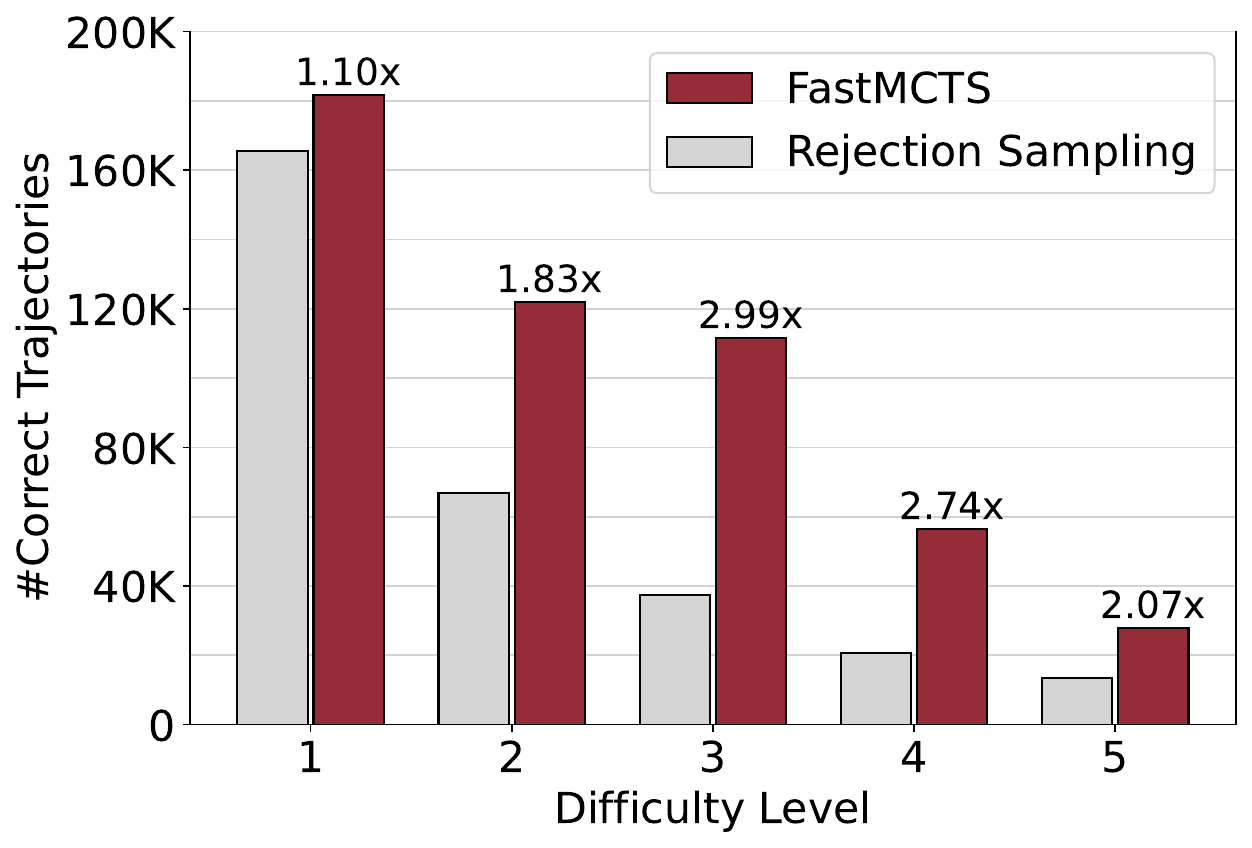}
        \caption{Sampling Balance on CN High School Math Hard}
        \label{fig:efficiency_cn}
    \end{subfigure}
    
    \caption{Comparison of sampling balance across difficulty levels for Rejection Sampling and FastMCTS.}
    \label{fig:balance}
\end{figure*}

\subsubsection{Difficulty-Aware Sampling in FastMCTS}
\label{sec:diffifulty}

\begin{table}[t]
\centering
\footnotesize
\renewcommand{\arraystretch}{1.4}
\begin{tabular}{lcc}
\hline
Method & EN Math Hard & \begin{tabular}[c]{@{}c@{}}CN High School\\  Math Hard \end{tabular}\\
\hline
Rejection Sampling & 2.10 & 1.79 \\
FastMCTS & \textbf{2.23} & \textbf{2.10} \\
\hline
\end{tabular}
\caption{The entropy comparison of difficulty level distributions (see Figure \ref{fig:balance}) in data synthesis methods.}
\label{tab:entropy}
\end{table}

As described in Section~\ref{sec:adaptive}, FastMCTS dynamically adapts the search process according to the problem difficulty. This adaptation results in a more balanced distribution of problems across different difficulty levels. Consequently, the data generated by FastMCTS is not only larger in quantity but also more effective for training purposes.

To analyze this, we categorize problems from our dataset into five difficulty levels based on the probability of sampling a correct answer using rejection sampling. We then compare the number of correct trajectories generated by both FastMCTS and rejection sampling for each level.

The results in Figure~\ref{fig:balance} show that FastMCTS achieves a more balanced distribution across difficulty levels than rejection sampling, particularly for higher-difficulty problems. These results highlight FastMCTS's difficulty-aware feature. During tree search, as iterations increase, Monte Carlo-estimated scores become more accurate. For harder problems, FastMCTS tends to sample branches with higher success probabilities, while for easier problems, it degenerates to rejection sampling, mainly focusing on diversity.

In Table \ref{tab:entropy}, we also report the entropy of the distribution presented in Figure~\ref{fig:balance}, which serves as a quantitative measure of its uniformity. The data synthesized by FastMCTS exhibits a higher entropy value, indicating a more uniform distribution across difficulty tiers compared to Rejection Sampling.

These findings explain the effectiveness of data synthesized by FastMCTS. Although tree-search process may reduce diversity due to shared prefixes, FastMCTS achieves a more balanced distribution across problems of varying difficulty levels.

\subsubsection{Ablation Study}
\label{sec:ablation}

\begin{table}[t]
\footnotesize
\begin{tabular}{lcc}
\toprule
Method                          & Solving Rate(\%)    & \#Correct Path  \\
\midrule
Rejection Sampling              & 61.3                   & 7.22                       \\
\midrule
FastMCTS     & \textbf{61.7}             & \textbf{7.95}       \\
\quad w/o fewshot   & 60.7             & 7.37             \\
\quad w/o stay   & 55.9             & 7.59             \\
\quad w/o dynamic   & \textbf{61.7}             & 7.28                           \\
\quad w/o stay \& dynamic & 55.9    & 7.32                     \\
\bottomrule
\end{tabular}
\caption{Ablation study}
\label{tab:ablation}
\end{table}

For our ablation study, we compare the efficiency of FastMCTS with and without Adaptive Stay and Dynamic Exploration, using Rejection Sampling as the baseline. Experiments are conducted on 300 randomly selected AIME problems under the same settings provided in Appendix~\ref{apd:sampling settings}. For each problem, we sample 25 trajectories: Rejection Sampling directly generates 25 trajectories, while FastMCTS performs 12 iterations of tree search with an initial degree of 3 and then expands 2 branches per phase, also yielding 25 trajectories.

From results in Table~\ref{tab:ablation} (averaged over multiple runs), we could deduce that
the Adaptive Stay policy primarily affects the problem solving rate. It decides whether to continue searching deeper or expand new branches based on the current node's score. As for Dynamic Exploration, it increases the efficiency of generating correct trajectories, as its absence reduces the average number of correct paths from 7.95 to 7.28. Removing few-shot examples leads to declines both in Problem Solving Rate and Average Correct Paths. These findings highlight the necessity and effectiveness of our proposed improvements in FastMCTS.

% These findings highlight the necessity and effectiveness of our proposed improvements. Together, these components enable FastMCTS to achieve superior performance in synthesizing multi-step reasoning data.

% These findings highlight the necessity and effectiveness of our proposed improvements. Together, these components enable FastMCTS to achieve superior performance in synthesizing multi-step reasoning data.

% 如果需要篇幅，就在这里提一下，this 证说明了我们方法改进之处的有效性之类的话

% \usepackage{multirow}

% 改成单栏 去掉那两列结果不必要的

% 是否可以再放一个难题目的表？

% 定义注释的方法

\section{Conclusion}

In this work, we introduce FastMCTS, an efficient sampling algorithm that leverages Monte Carlo Tree Search to synthesize high-quality multi-step reasoning data for training large language models. Our approach not only improves the efficiency of data synthesis but also promotes a balanced sampling distribution across problems of varying difficulty, while providing step-level supervision for enhanced training like DPO. Experimental results demonstrate that FastMCTS outperforms rejection sampling in both sampling efficiency and training performance under comparable synthetic data budgets. We believe our method offers a practical solution for efficiently generating high-quality multi-step reasoning data and hope it inspires further research on data synthesis for language models.

% \newpage
% without newpage!

\section*{Limitations}
Our work has several limitations. First, although we utilize a diverse range of data sources for data synthesis, our synthetic data is generated solely by the open-source model Qwen2.5-72B-Instruct for data generation. We do not employ stronger closed-source models like GPT-4 or models specifically fine-tuned for higher reasoning capabilities, such as Qwen-Math~\cite{DBLP:journals/corr/abs-2409-12122}, DeepSeek-R1~\cite{deepseekai2025deepseekr1incentivizingreasoningcapability}, or o1~\cite{openai2024openaio1card}. As a result, the performance of the trained models is not state-of-the-art. 

Additionally, due to computational budget, we conduct our synthetic data experiments only in the math domain, we plan to extend our experiments to data from other domains in future work.

Finally, while FastMCTS-synthesized data achieve better training results due to its quantity and balanced distribution, the impact of prefix repetition in reasoning paths caused by the tree structure remains an open question, which we plan to investigate in future work.


% \clearpage

\section*{Acknowledgments}
This work was supported by the National Key Research and Development Program of China (No. 2022ZD0160102). We express our sincere gratitude to Fudan University and Shanghai Artificial Intelligence Laboratory for providing an outstanding research environment, invaluable resources, and continuous support throughout this work.

\bibliography{custom}

\begin{thebibliography}{50}
\providecommand{\natexlab}[1]{#1}

\bibitem[{Achiam et~al.(2023)Achiam, Adler, Agarwal, Ahmad, Akkaya, Aleman, Almeida, Altenschmidt, Altman, Anadkat, Avila, Babuschkin, Balaji, Balcom, Baltescu, ing Bao, Bavarian, Belgum, Bello, Berdine, Bernadett-Shapiro, Berner, Bogdonoff, Boiko, Boyd, Brakman, Brockman, Brooks, Brundage, Button, Cai, Campbell, Cann, Carey, Carlson, Carmichael, Chan, Chang, Chantzis, Chen, Chen, Chen, Chen, Chen, Chess, Cho, Chu, Chung, Cummings, Currier, Dai, Decareaux, Degry, Deutsch, Deville, Dhar, Dohan, Dowling, Dunning, Ecoffet, Eleti, Eloundou, Farhi, Fedus, Felix, Fishman, Forte, abella Fulford, Gao, Georges, Gibson, Goel, Gogineni, Goh, Gontijo-Lopes, Gordon, Grafstein, Gray, Greene, Gross, Gu, Guo, Hallacy, Han, Harris, He, Heaton, Heidecke, Hesse, Hickey, Hickey, Hoeschele, Houghton, Hsu, Hu, Hu, Huizinga, Jain, Jain, Jang, Jiang, Jiang, Jin, Jin, Jomoto, Jonn, Jun, Kaftan, Kaiser, Kamali, Kanitscheider, Keskar, Khan, Kilpatrick, Kim, Kim, Kim, Kirchner, Kiros, Knight, Kokotajlo, Kondraciuk, Kondrich,
  Konstantinidis, Kosic, Krueger, Kuo, Lampe, Lan, Lee, Leike, Leung, Levy, Li, Lim, Lin, Lin, teusz Litwin, Lopez, Lowe, Lue, Makanju, Malfacini, Manning, Markov, Markovski, Martin, Mayer, Mayne, McGrew, McKinney, McLeavey, McMillan, McNeil, Medina, Mehta, Menick, Metz, Mishchenko, Mishkin, Monaco, Morikawa, Mossing, Mu, Murati, Murk, M'ely, Nair, Nakano, Nayak, Neelakantan, Ngo, Noh, Long, O'Keefe, Pachocki, Paino, Palermo, Pantuliano, Parascandolo, Parish, Parparita, Passos, Pavlov, Peng, Perelman, de~Avila Belbute~Peres, Petrov, de~Oliveira~Pinto, Pokorny, Pokrass, Pong, Powell, Power, Power, Proehl, Puri, Radford, Rae, Ramesh, Raymond, Real, Rimbach, Ross, Rotsted, Roussez, Ryder, Saltarelli, Sanders, Santurkar, Sastry, Schmidt, Schnurr, Schulman, Selsam, Sheppard, Sherbakov, Shieh, Shoker, Shyam, Sidor, Sigler, Simens, Sitkin, Slama, Sohl, Sokolowsky, Song, Staudacher, Such, Summers, Sutskever, Tang, Tezak, Thompson, Tillet, Tootoonchian, Tseng, Tuggle, Turley, Tworek, Uribe, Vallone, Vijayvergiya,
  Voss, Wainwright, Wang, Wang, Wang, Ward, Wei, Weinmann, Welihinda, Welinder, Weng, Weng, Wiethoff, Willner, Winter, Wolrich, Wong, Workman, Wu, Wu, Wu, Xiao, Xu, Yoo, Yu, ing Yuan, Zaremba, Zellers, Zhang, Zhang, Zhao, Zheng, Zhuang, Zhuk, and Zoph}]{Achiam2023GPT4TR}
OpenAI~Josh Achiam, Steven Adler, Sandhini Agarwal, Lama Ahmad, Ilge Akkaya, Florencia~Leoni Aleman, Diogo Almeida, Janko Altenschmidt, Sam Altman, Shyamal Anadkat, Red Avila, Igor Babuschkin, Suchir Balaji, Valerie Balcom, Paul Baltescu, Haim ing Bao, Mo~Bavarian, Jeff Belgum, Irwan Bello, Jake Berdine, Gabriel Bernadett-Shapiro, Christopher Berner, Lenny Bogdonoff, Oleg Boiko, Madelaine Boyd, Anna-Luisa Brakman, Greg Brockman, Tim Brooks, Miles Brundage, Kevin Button, Trevor Cai, Rosie Campbell, Andrew Cann, Brittany Carey, Chelsea Carlson, Rory Carmichael, Brooke Chan, Che Chang, Fotis Chantzis, Derek Chen, Sully Chen, Ruby Chen, Jason Chen, Mark Chen, Benjamin Chess, Chester Cho, Casey Chu, Hyung~Won Chung, Dave Cummings, Jeremiah Currier, Yunxing Dai, Cory Decareaux, Thomas Degry, Noah Deutsch, Damien Deville, Arka Dhar, David Dohan, Steve Dowling, Sheila Dunning, Adrien Ecoffet, Atty Eleti, Tyna Eloundou, David Farhi, Liam Fedus, Niko Felix, Sim'on~Posada Fishman, Juston Forte, Is~abella Fulford, Leo
  Gao, Elie Georges, Christian Gibson, Vik Goel, Tarun Gogineni, Gabriel Goh, Raphael Gontijo-Lopes, Jonathan Gordon, Morgan Grafstein, Scott Gray, Ryan Greene, Joshua Gross, Shixiang~Shane Gu, Yufei Guo, Chris Hallacy, Jesse Han, Jeff Harris, Yuchen He, Mike Heaton, Johannes Heidecke, Chris Hesse, Alan Hickey, Wade Hickey, Peter Hoeschele, Brandon Houghton, Kenny Hsu, Shengli Hu, Xin Hu, Joost Huizinga, Shantanu Jain, Shawn Jain, Joanne Jang, Angela Jiang, Roger Jiang, Haozhun Jin, Denny Jin, Shino Jomoto, Billie Jonn, Heewoo Jun, Tomer Kaftan, Lukasz Kaiser, Ali Kamali, Ingmar Kanitscheider, Nitish~Shirish Keskar, Tabarak Khan, Logan Kilpatrick, Jong~Wook Kim, Christina Kim, Yongjik Kim, Hendrik Kirchner, Jamie~Ryan Kiros, Matthew Knight, Daniel Kokotajlo, Lukasz Kondraciuk, Andrew Kondrich, Aris Konstantinidis, Kyle Kosic, Gretchen Krueger, Vishal Kuo, Michael Lampe, Ikai Lan, Teddy Lee, Jan Leike, Jade Leung, Daniel Levy, Chak~Ming Li, Rachel Lim, Molly Lin, Stephanie Lin, Ma~teusz Litwin, Theresa Lopez,
  Ryan Lowe, Patricia Lue, Anna Makanju, Kim Malfacini, Sam Manning, Todor Markov, Yaniv Markovski, Bianca Martin, Katie Mayer, Andrew Mayne, Bob McGrew, Scott~Mayer McKinney, Christine McLeavey, Paul McMillan, Jake McNeil, David Medina, Aalok Mehta, Jacob Menick, Luke Metz, Andrey Mishchenko, Pamela Mishkin, Vinnie Monaco, Evan Morikawa, Daniel~P. Mossing, Tong Mu, Mira Murati, Oleg Murk, David M'ely, Ashvin Nair, Reiichiro Nakano, Rajeev Nayak, Arvind Neelakantan, Richard Ngo, Hyeonwoo Noh, Ouyang Long, Cullen O'Keefe, Jakub~W. Pachocki, Alex Paino, Joe Palermo, Ashley Pantuliano, Giambattista Parascandolo, Joel Parish, Emy Parparita, Alexandre Passos, Mikhail Pavlov, Andrew Peng, Adam Perelman, Filipe de~Avila Belbute~Peres, Michael Petrov, Henrique~Pond{\'e} de~Oliveira~Pinto, Michael Pokorny, Michelle Pokrass, Vitchyr~H. Pong, Tolly Powell, Alethea Power, Boris Power, Elizabeth Proehl, Raul Puri, Alec Radford, Jack~W. Rae, Aditya Ramesh, Cameron Raymond, Francis Real, Kendra Rimbach, Carl Ross, Bob
  Rotsted, Henri Roussez, Nick Ryder, Mario~D. Saltarelli, Ted Sanders, Shibani Santurkar, Girish Sastry, Heather Schmidt, David Schnurr, John Schulman, Daniel Selsam, Kyla Sheppard, Toki Sherbakov, Jessica Shieh, Sarah Shoker, Pranav Shyam, Szymon Sidor, Eric Sigler, Maddie Simens, Jordan Sitkin, Katarina Slama, Ian Sohl, Benjamin~D. Sokolowsky, Yang Song, Natalie Staudacher, Felipe~Petroski Such, Natalie Summers, Ilya Sutskever, Jie Tang, Nikolas~A. Tezak, Madeleine Thompson, Phil Tillet, Amin Tootoonchian, Elizabeth Tseng, Preston Tuggle, Nick Turley, Jerry Tworek, Juan Felipe~Cer'on Uribe, Andrea Vallone, Arun Vijayvergiya, Chelsea Voss, Carroll~L. Wainwright, Justin~Jay Wang, Alvin Wang, Ben Wang, Jonathan Ward, Jason Wei, CJ~Weinmann, Akila Welihinda, Peter Welinder, Jiayi Weng, Lilian Weng, Matt Wiethoff, Dave Willner, Clemens Winter, Samuel Wolrich, Hannah Wong, Lauren Workman, Sherwin Wu, Jeff Wu, Michael Wu, Kai Xiao, Tao Xu, Sarah Yoo, Kevin Yu, Qim ing Yuan, Wojciech Zaremba, Rowan Zellers, Chong
  Zhang, Marvin Zhang, Shengjia Zhao, Tianhao Zheng, Juntang Zhuang, William Zhuk, and Barret Zoph. 2023.
\newblock \href {https://api.semanticscholar.org/CorpusID:257532815} {Gpt-4 technical report}.

\bibitem[{AI-MO(2023{\natexlab{a}})}]{aime}
AI-MO. 2023{\natexlab{a}}.
\newblock Aime problems and solutions.
\newblock \url{https://artofproblemsolving.com/wiki/index.php/AIME_Problems_and_Solutions}.

\bibitem[{AI-MO(2023{\natexlab{b}})}]{amc}
AI-MO. 2023{\natexlab{b}}.
\newblock American mathematics competitions.
\newblock \url{https://artofproblemsolving.com/wiki/index.php/AMC_12_Problems_and_Solutions}.

\bibitem[{AI-MO(2024)}]{aime24}
AI-MO. 2024.
\newblock Aime problems and solutions.
\newblock \url{https://artofproblemsolving.com/wiki/index.php/AIME_Problems_and_Solutions}.

\bibitem[{Browne et~al.(2012)Browne, Powley, Whitehouse, Lucas, Cowling, Rohlfshagen, Tavener, Liebana, Samothrakis, and Colton}]{DBLP:journals/tciaig/BrownePWLCRTPSC12}
Cameron Browne, Edward~Jack Powley, Daniel Whitehouse, Simon~M. Lucas, Peter~I. Cowling, Philipp Rohlfshagen, Stephen Tavener, Diego~Perez Liebana, Spyridon Samothrakis, and Simon Colton. 2012.
\newblock \href {https://doi.org/10.1109/TCIAIG.2012.2186810} {A survey of monte carlo tree search methods}.
\newblock \emph{{IEEE} Trans. Comput. Intell. {AI} Games}, 4(1):1--43.

\bibitem[{Chen et~al.(2024)Chen, Liao, Li, and Fan}]{DBLP:journals/corr/abs-2405-03553}
Guoxin Chen, Minpeng Liao, Chengxi Li, and Kai Fan. 2024.
\newblock \href {https://doi.org/10.48550/ARXIV.2405.03553} {Alphamath almost zero: process supervision without process}.
\newblock \emph{CoRR}, abs/2405.03553.

\bibitem[{Cobbe et~al.(2021)Cobbe, Kosaraju, Bavarian, Chen, Jun, Kaiser, Plappert, Tworek, Hilton, Nakano, Hesse, and Schulman}]{DBLP:journals/corr/abs-2110-14168}
Karl Cobbe, Vineet Kosaraju, Mohammad Bavarian, Mark Chen, Heewoo Jun, Lukasz Kaiser, Matthias Plappert, Jerry Tworek, Jacob Hilton, Reiichiro Nakano, Christopher Hesse, and John Schulman. 2021.
\newblock \href {https://arxiv.org/abs/2110.14168} {Training verifiers to solve math word problems}.
\newblock \emph{CoRR}, abs/2110.14168.

\bibitem[{Coulom(2006)}]{DBLP:conf/cg/Coulom06}
R{\'{e}}mi Coulom. 2006.
\newblock \href {https://doi.org/10.1007/978-3-540-75538-8\_7} {Efficient selectivity and backup operators in monte-carlo tree search}.
\newblock In \emph{Computers and Games, 5th International Conference, {CG} 2006, Turin, Italy, May 29-31, 2006. Revised Papers}, volume 4630 of \emph{Lecture Notes in Computer Science}, pages 72--83. Springer.

\bibitem[{DeepSeek-AI et~al.(2025)DeepSeek-AI, Guo, Yang, Zhang, Song, Zhang, Xu, Zhu, Ma, Wang, Bi, Zhang, Yu, Wu, Wu, Gou, Shao, Li, Gao, Liu, Xue, Wang, Wu, Feng, Lu, Zhao, Deng, Zhang, Ruan, Dai, Chen, Ji, Li, Lin, Dai, Luo, Hao, Chen, Li, Zhang, Bao, Xu, Wang, Ding, Xin, Gao, Qu, Li, Guo, Li, Wang, Chen, Yuan, Qiu, Li, Cai, Ni, Liang, Chen, Dong, Hu, Gao, Guan, Huang, Yu, Wang, Zhang, Zhao, Wang, Zhang, Xu, Xia, Zhang, Zhang, Tang, Li, Wang, Li, Tian, Huang, Zhang, Wang, Chen, Du, Ge, Zhang, Pan, Wang, Chen, Jin, Chen, Lu, Zhou, Chen, Ye, Wang, Yu, Zhou, Pan, Li, Zhou, Wu, Ye, Yun, Pei, Sun, Wang, Zeng, Zhao, Liu, Liang, Gao, Yu, Zhang, Xiao, An, Liu, Wang, Chen, Nie, Cheng, Liu, Xie, Liu, Yang, Li, Su, Lin, Li, Jin, Shen, Chen, Sun, Wang, Song, Zhou, Wang, Shan, Li, Wang, Wei, Zhang, Xu, Li, Zhao, Sun, Wang, Yu, Zhang, Shi, Xiong, He, Piao, Wang, Tan, Ma, Liu, Guo, Ou, Wang, Gong, Zou, He, Xiong, Luo, You, Liu, Zhou, Zhu, Xu, Huang, Li, Zheng, Zhu, Ma, Tang, Zha, Yan, Ren, Ren, Sha, Fu, Xu, Xie, Zhang,
  Hao, Ma, Yan, Wu, Gu, Zhu, Liu, Li, Xie, Song, Pan, Huang, Xu, Zhang, and Zhang}]{deepseekai2025deepseekr1incentivizingreasoningcapability}
DeepSeek-AI, Daya Guo, Dejian Yang, Haowei Zhang, Junxiao Song, Ruoyu Zhang, Runxin Xu, Qihao Zhu, Shirong Ma, Peiyi Wang, Xiao Bi, Xiaokang Zhang, Xingkai Yu, Yu~Wu, Z.~F. Wu, Zhibin Gou, Zhihong Shao, Zhuoshu Li, Ziyi Gao, Aixin Liu, Bing Xue, Bingxuan Wang, Bochao Wu, Bei Feng, Chengda Lu, Chenggang Zhao, Chengqi Deng, Chenyu Zhang, Chong Ruan, Damai Dai, Deli Chen, Dongjie Ji, Erhang Li, Fangyun Lin, Fucong Dai, Fuli Luo, Guangbo Hao, Guanting Chen, Guowei Li, H.~Zhang, Han Bao, Hanwei Xu, Haocheng Wang, Honghui Ding, Huajian Xin, Huazuo Gao, Hui Qu, Hui Li, Jianzhong Guo, Jiashi Li, Jiawei Wang, Jingchang Chen, Jingyang Yuan, Junjie Qiu, Junlong Li, J.~L. Cai, Jiaqi Ni, Jian Liang, Jin Chen, Kai Dong, Kai Hu, Kaige Gao, Kang Guan, Kexin Huang, Kuai Yu, Lean Wang, Lecong Zhang, Liang Zhao, Litong Wang, Liyue Zhang, Lei Xu, Leyi Xia, Mingchuan Zhang, Minghua Zhang, Minghui Tang, Meng Li, Miaojun Wang, Mingming Li, Ning Tian, Panpan Huang, Peng Zhang, Qiancheng Wang, Qinyu Chen, Qiushi Du, Ruiqi Ge, Ruisong
  Zhang, Ruizhe Pan, Runji Wang, R.~J. Chen, R.~L. Jin, Ruyi Chen, Shanghao Lu, Shangyan Zhou, Shanhuang Chen, Shengfeng Ye, Shiyu Wang, Shuiping Yu, Shunfeng Zhou, Shuting Pan, S.~S. Li, Shuang Zhou, Shaoqing Wu, Shengfeng Ye, Tao Yun, Tian Pei, Tianyu Sun, T.~Wang, Wangding Zeng, Wanjia Zhao, Wen Liu, Wenfeng Liang, Wenjun Gao, Wenqin Yu, Wentao Zhang, W.~L. Xiao, Wei An, Xiaodong Liu, Xiaohan Wang, Xiaokang Chen, Xiaotao Nie, Xin Cheng, Xin Liu, Xin Xie, Xingchao Liu, Xinyu Yang, Xinyuan Li, Xuecheng Su, Xuheng Lin, X.~Q. Li, Xiangyue Jin, Xiaojin Shen, Xiaosha Chen, Xiaowen Sun, Xiaoxiang Wang, Xinnan Song, Xinyi Zhou, Xianzu Wang, Xinxia Shan, Y.~K. Li, Y.~Q. Wang, Y.~X. Wei, Yang Zhang, Yanhong Xu, Yao Li, Yao Zhao, Yaofeng Sun, Yaohui Wang, Yi~Yu, Yichao Zhang, Yifan Shi, Yiliang Xiong, Ying He, Yishi Piao, Yisong Wang, Yixuan Tan, Yiyang Ma, Yiyuan Liu, Yongqiang Guo, Yuan Ou, Yuduan Wang, Yue Gong, Yuheng Zou, Yujia He, Yunfan Xiong, Yuxiang Luo, Yuxiang You, Yuxuan Liu, Yuyang Zhou, Y.~X. Zhu,
  Yanhong Xu, Yanping Huang, Yaohui Li, Yi~Zheng, Yuchen Zhu, Yunxian Ma, Ying Tang, Yukun Zha, Yuting Yan, Z.~Z. Ren, Zehui Ren, Zhangli Sha, Zhe Fu, Zhean Xu, Zhenda Xie, Zhengyan Zhang, Zhewen Hao, Zhicheng Ma, Zhigang Yan, Zhiyu Wu, Zihui Gu, Zijia Zhu, Zijun Liu, Zilin Li, Ziwei Xie, Ziyang Song, Zizheng Pan, Zhen Huang, Zhipeng Xu, Zhongyu Zhang, and Zhen Zhang. 2025.
\newblock \href {https://arxiv.org/abs/2501.12948} {Deepseek-r1: Incentivizing reasoning capability in llms via reinforcement learning}.
\newblock \emph{Preprint}, arXiv:2501.12948.

\bibitem[{Feng et~al.(2023)Feng, Wan, Wen, McAleer, Wen, Zhang, and Wang}]{feng2023alphazero}
Xidong Feng, Ziyu Wan, Muning Wen, Stephen~Marcus McAleer, Ying Wen, Weinan Zhang, and Jun Wang. 2023.
\newblock Alphazero-like tree-search can guide large language model decoding and training.
\newblock \emph{arXiv preprint arXiv:2309.17179}.

\bibitem[{Gao et~al.(2024)Gao, Song, Yang, Cai, Miao, Dong, Li, Ma, Chen, Xu, Tang, Wang, Zan, Quan, Zhang, Sha, Zhang, Ren, Liu, and Chang}]{DBLP:journals/corr/abs-2410-07985}
Bofei Gao, Feifan Song, Zhe Yang, Zefan Cai, Yibo Miao, Qingxiu Dong, Lei Li, Chenghao Ma, Liang Chen, Runxin Xu, Zhengyang Tang, Benyou Wang, Daoguang Zan, Shanghaoran Quan, Ge~Zhang, Lei Sha, Yichang Zhang, Xuancheng Ren, Tianyu Liu, and Baobao Chang. 2024.
\newblock \href {https://doi.org/10.48550/ARXIV.2410.07985} {Omni-math: {A} universal olympiad level mathematic benchmark for large language models}.
\newblock \emph{CoRR}, abs/2410.07985.

\bibitem[{Hao et~al.(2023)Hao, Gu, Ma, Hong, Wang, Wang, and Hu}]{hao2023reasoning}
Shibo Hao, Yi~Gu, Haodi Ma, Joshua~Jiahua Hong, Zhen Wang, Daisy~Zhe Wang, and Zhiting Hu. 2023.
\newblock Reasoning with language model is planning with world model.
\newblock \emph{arXiv preprint arXiv:2305.14992}.

\bibitem[{He et~al.(2024)He, Luo, Bai, Hu, Thai, Shen, Hu, Han, Huang, Zhang, Liu, Qi, Liu, and Sun}]{DBLP:conf/acl/HeLBHTSHHHZLQL024}
Chaoqun He, Renjie Luo, Yuzhuo Bai, Shengding Hu, Zhen~Leng Thai, Junhao Shen, Jinyi Hu, Xu~Han, Yujie Huang, Yuxiang Zhang, Jie Liu, Lei Qi, Zhiyuan Liu, and Maosong Sun. 2024.
\newblock \href {https://doi.org/10.18653/V1/2024.ACL-LONG.211} {Olympiadbench: {A} challenging benchmark for promoting {AGI} with olympiad-level bilingual multimodal scientific problems}.
\newblock In \emph{Proceedings of the 62nd Annual Meeting of the Association for Computational Linguistics (Volume 1: Long Papers), {ACL} 2024, Bangkok, Thailand, August 11-16, 2024}, pages 3828--3850. Association for Computational Linguistics.

\bibitem[{Hendrycks et~al.(2021)Hendrycks, Burns, Kadavath, Arora, Basart, Tang, Song, and Steinhardt}]{hendrycksmath2021}
Dan Hendrycks, Collin Burns, Saurav Kadavath, Akul Arora, Steven Basart, Eric Tang, Dawn Song, and Jacob Steinhardt. 2021.
\newblock Measuring mathematical problem solving with the math dataset.
\newblock \emph{NeurIPS}.

\bibitem[{Lai et~al.(2024)Lai, Tian, Chen, Yang, Peng, and Jia}]{lai2024stepdpostepwisepreferenceoptimization}
Xin Lai, Zhuotao Tian, Yukang Chen, Senqiao Yang, Xiangru Peng, and Jiaya Jia. 2024.
\newblock \href {https://arxiv.org/abs/2406.18629} {Step-dpo: Step-wise preference optimization for long-chain reasoning of llms}.
\newblock \emph{Preprint}, arXiv:2406.18629.

\bibitem[{Li et~al.(2024)Li, Wang, Hu, Wei, Zheng, Hu, Zhang, and Peng}]{DBLP:journals/corr/abs-2403-04706}
Chen Li, Weiqi Wang, Jingcheng Hu, Yixuan Wei, Nanning Zheng, Han Hu, Zheng Zhang, and Houwen Peng. 2024.
\newblock \href {https://doi.org/10.48550/ARXIV.2403.04706} {Common 7b language models already possess strong math capabilities}.
\newblock \emph{CoRR}, abs/2403.04706.

\bibitem[{LI et~al.(2024)LI, Beeching, Tunstall, Lipkin, Soletskyi, Huang, Rasul, Yu, Jiang, Shen, Qin, Dong, Zhou, Fleureau, Lample, and Polu}]{numina_math_datasets}
Jia LI, Edward Beeching, Lewis Tunstall, Ben Lipkin, Roman Soletskyi, Shengyi~Costa Huang, Kashif Rasul, Longhui Yu, Albert Jiang, Ziju Shen, Zihan Qin, Bin Dong, Li~Zhou, Yann Fleureau, Guillaume Lample, and Stanislas Polu. 2024.
\newblock Numinamath.
\newblock \url{[https://huggingface.co/AI-MO/NuminaMath-CoT](https://github.com/project-numina/aimo-progress-prize/blob/main/report/numina_dataset.pdf)}.

\bibitem[{Lightman et~al.(2024)Lightman, Kosaraju, Burda, Edwards, Baker, Lee, Leike, Schulman, Sutskever, and Cobbe}]{DBLP:conf/iclr/LightmanKBEBLLS24}
Hunter Lightman, Vineet Kosaraju, Yuri Burda, Harrison Edwards, Bowen Baker, Teddy Lee, Jan Leike, John Schulman, Ilya Sutskever, and Karl Cobbe. 2024.
\newblock \href {https://openreview.net/forum?id=v8L0pN6EOi} {Let's verify step by step}.
\newblock In \emph{The Twelfth International Conference on Learning Representations, {ICLR} 2024, Vienna, Austria, May 7-11, 2024}. OpenReview.net.

\bibitem[{Luo et~al.(2024)Luo, Liu, Liu, Phatale, Lara, Li, Shu, Zhu, Meng, Sun, and Rastogi}]{DBLP:journals/corr/abs-2406-06592}
Liangchen Luo, Yinxiao Liu, Rosanne Liu, Samrat Phatale, Harsh Lara, Yunxuan Li, Lei Shu, Yun Zhu, Lei Meng, Jiao Sun, and Abhinav Rastogi. 2024.
\newblock \href {https://doi.org/10.48550/ARXIV.2406.06592} {Improve mathematical reasoning in language models by automated process supervision}.
\newblock \emph{CoRR}, abs/2406.06592.

\bibitem[{Mukherjee et~al.(2023)Mukherjee, Mitra, Jawahar, Agarwal, Palangi, and Awadallah}]{DBLP:journals/corr/abs-2306-02707}
Subhabrata Mukherjee, Arindam Mitra, Ganesh Jawahar, Sahaj Agarwal, Hamid Palangi, and Ahmed Awadallah. 2023.
\newblock \href {https://doi.org/10.48550/ARXIV.2306.02707} {Orca: Progressive learning from complex explanation traces of {GPT-4}}.
\newblock \emph{CoRR}, abs/2306.02707.

\bibitem[{Neal(2003)}]{neal2003slice}
Radford~M Neal. 2003.
\newblock Slice sampling.
\newblock \emph{The annals of statistics}, 31(3):705--767.

\bibitem[{OpenAI et~al.(2024)OpenAI, :, Jaech, Kalai, Lerer, Richardson, El-Kishky, Low, Helyar, Madry, Beutel, Carney, Iftimie, Karpenko, Passos, Neitz, Prokofiev, Wei, Tam, Bennett, Kumar, Saraiva, Vallone, Duberstein, Kondrich, Mishchenko, Applebaum, Jiang, Nair, Zoph, Ghorbani, Rossen, Sokolowsky, Barak, McGrew, Minaiev, Hao, Baker, Houghton, McKinzie, Eastman, Lugaresi, Bassin, Hudson, Li, de~Bourcy, Voss, Shen, Zhang, Koch, Orsinger, Hesse, Fischer, Chan, Roberts, Kappler, Levy, Selsam, Dohan, Farhi, Mely, Robinson, Tsipras, Li, Oprica, Freeman, Zhang, Wong, Proehl, Cheung, Mitchell, Wallace, Ritter, Mays, Wang, Such, Raso, Leoni, Tsimpourlas, Song, von Lohmann, Sulit, Salmon, Parascandolo, Chabot, Zhao, Brockman, Leclerc, Salman, Bao, Sheng, Andrin, Bagherinezhad, Ren, Lightman, Chung, Kivlichan, O'Connell, Osband, Gilaberte, Akkaya, Kostrikov, Sutskever, Kofman, Pachocki, Lennon, Wei, Harb, Twore, Feng, Yu, Weng, Tang, Yu, Candela, Palermo, Parish, Heidecke, Hallman, Rizzo, Gordon, Uesato, Ward,
  Huizinga, Wang, Chen, Xiao, Singhal, Nguyen, Cobbe, Shi, Wood, Rimbach, Gu-Lemberg, Liu, Lu, Stone, Yu, Ahmad, Yang, Liu, Maksin, Ho, Fedus, Weng, Li, McCallum, Held, Kuhn, Kondraciuk, Kaiser, Metz, Boyd, Trebacz, Joglekar, Chen, Tintor, Meyer, Jones, Kaufer, Schwarzer, Shah, Yatbaz, Guan, Xu, Yan, Glaese, Chen, Lampe, Malek, Wang, Fradin, McClay, Pavlov, Wang, Wang, Murati, Bavarian, Rohaninejad, McAleese, Chowdhury, Chowdhury, Ryder, Tezak, Brown, Nachum, Boiko, Murk, Watkins, Chao, Ashbourne, Izmailov, Zhokhov, Dias, Arora, Lin, Lopes, Gaon, Miyara, Leike, Hwang, Garg, Brown, James, Shu, Cheu, Greene, Jain, Altman, Toizer, Toyer, Miserendino, Agarwal, Hernandez, Baker, McKinney, Yan, Zhao, Hu, Santurkar, Chaudhuri, Zhang, Fu, Papay, Lin, Balaji, Sanjeev, Sidor, Broda, Clark, Wang, Gordon, Sanders, Patwardhan, Sottiaux, Degry, Dimson, Zheng, Garipov, Stasi, Bansal, Creech, Peterson, Eloundou, Qi, Kosaraju, Monaco, Pong, Fomenko, Zheng, Zhou, McCabe, Zaremba, Dubois, Lu, Chen, Cha, Bai, He, Zhang, Wang,
  Shao, and Li}]{openai2024openaio1card}
OpenAI, :, Aaron Jaech, Adam Kalai, Adam Lerer, Adam Richardson, Ahmed El-Kishky, Aiden Low, Alec Helyar, Aleksander Madry, Alex Beutel, Alex Carney, Alex Iftimie, Alex Karpenko, Alex~Tachard Passos, Alexander Neitz, Alexander Prokofiev, Alexander Wei, Allison Tam, Ally Bennett, Ananya Kumar, Andre Saraiva, Andrea Vallone, Andrew Duberstein, Andrew Kondrich, Andrey Mishchenko, Andy Applebaum, Angela Jiang, Ashvin Nair, Barret Zoph, Behrooz Ghorbani, Ben Rossen, Benjamin Sokolowsky, Boaz Barak, Bob McGrew, Borys Minaiev, Botao Hao, Bowen Baker, Brandon Houghton, Brandon McKinzie, Brydon Eastman, Camillo Lugaresi, Cary Bassin, Cary Hudson, Chak~Ming Li, Charles de~Bourcy, Chelsea Voss, Chen Shen, Chong Zhang, Chris Koch, Chris Orsinger, Christopher Hesse, Claudia Fischer, Clive Chan, Dan Roberts, Daniel Kappler, Daniel Levy, Daniel Selsam, David Dohan, David Farhi, David Mely, David Robinson, Dimitris Tsipras, Doug Li, Dragos Oprica, Eben Freeman, Eddie Zhang, Edmund Wong, Elizabeth Proehl, Enoch Cheung, Eric
  Mitchell, Eric Wallace, Erik Ritter, Evan Mays, Fan Wang, Felipe~Petroski Such, Filippo Raso, Florencia Leoni, Foivos Tsimpourlas, Francis Song, Fred von Lohmann, Freddie Sulit, Geoff Salmon, Giambattista Parascandolo, Gildas Chabot, Grace Zhao, Greg Brockman, Guillaume Leclerc, Hadi Salman, Haiming Bao, Hao Sheng, Hart Andrin, Hessam Bagherinezhad, Hongyu Ren, Hunter Lightman, Hyung~Won Chung, Ian Kivlichan, Ian O'Connell, Ian Osband, Ignasi~Clavera Gilaberte, Ilge Akkaya, Ilya Kostrikov, Ilya Sutskever, Irina Kofman, Jakub Pachocki, James Lennon, Jason Wei, Jean Harb, Jerry Twore, Jiacheng Feng, Jiahui Yu, Jiayi Weng, Jie Tang, Jieqi Yu, Joaquin~Quiñonero Candela, Joe Palermo, Joel Parish, Johannes Heidecke, John Hallman, John Rizzo, Jonathan Gordon, Jonathan Uesato, Jonathan Ward, Joost Huizinga, Julie Wang, Kai Chen, Kai Xiao, Karan Singhal, Karina Nguyen, Karl Cobbe, Katy Shi, Kayla Wood, Kendra Rimbach, Keren Gu-Lemberg, Kevin Liu, Kevin Lu, Kevin Stone, Kevin Yu, Lama Ahmad, Lauren Yang, Leo Liu,
  Leon Maksin, Leyton Ho, Liam Fedus, Lilian Weng, Linden Li, Lindsay McCallum, Lindsey Held, Lorenz Kuhn, Lukas Kondraciuk, Lukasz Kaiser, Luke Metz, Madelaine Boyd, Maja Trebacz, Manas Joglekar, Mark Chen, Marko Tintor, Mason Meyer, Matt Jones, Matt Kaufer, Max Schwarzer, Meghan Shah, Mehmet Yatbaz, Melody~Y. Guan, Mengyuan Xu, Mengyuan Yan, Mia Glaese, Mianna Chen, Michael Lampe, Michael Malek, Michele Wang, Michelle Fradin, Mike McClay, Mikhail Pavlov, Miles Wang, Mingxuan Wang, Mira Murati, Mo~Bavarian, Mostafa Rohaninejad, Nat McAleese, Neil Chowdhury, Neil Chowdhury, Nick Ryder, Nikolas Tezak, Noam Brown, Ofir Nachum, Oleg Boiko, Oleg Murk, Olivia Watkins, Patrick Chao, Paul Ashbourne, Pavel Izmailov, Peter Zhokhov, Rachel Dias, Rahul Arora, Randall Lin, Rapha~Gontijo Lopes, Raz Gaon, Reah Miyara, Reimar Leike, Renny Hwang, Rhythm Garg, Robin Brown, Roshan James, Rui Shu, Ryan Cheu, Ryan Greene, Saachi Jain, Sam Altman, Sam Toizer, Sam Toyer, Samuel Miserendino, Sandhini Agarwal, Santiago Hernandez,
  Sasha Baker, Scott McKinney, Scottie Yan, Shengjia Zhao, Shengli Hu, Shibani Santurkar, Shraman~Ray Chaudhuri, Shuyuan Zhang, Siyuan Fu, Spencer Papay, Steph Lin, Suchir Balaji, Suvansh Sanjeev, Szymon Sidor, Tal Broda, Aidan Clark, Tao Wang, Taylor Gordon, Ted Sanders, Tejal Patwardhan, Thibault Sottiaux, Thomas Degry, Thomas Dimson, Tianhao Zheng, Timur Garipov, Tom Stasi, Trapit Bansal, Trevor Creech, Troy Peterson, Tyna Eloundou, Valerie Qi, Vineet Kosaraju, Vinnie Monaco, Vitchyr Pong, Vlad Fomenko, Weiyi Zheng, Wenda Zhou, Wes McCabe, Wojciech Zaremba, Yann Dubois, Yinghai Lu, Yining Chen, Young Cha, Yu~Bai, Yuchen He, Yuchen Zhang, Yunyun Wang, Zheng Shao, and Zhuohan Li. 2024.
\newblock \href {https://arxiv.org/abs/2412.16720} {Openai o1 system card}.
\newblock \emph{Preprint}, arXiv:2412.16720.

\bibitem[{Rafailov et~al.(2023)Rafailov, Sharma, Mitchell, Manning, Ermon, and Finn}]{DBLP:conf/nips/RafailovSMMEF23}
Rafael Rafailov, Archit Sharma, Eric Mitchell, Christopher~D. Manning, Stefano Ermon, and Chelsea Finn. 2023.
\newblock \href {http://papers.nips.cc/paper\_files/paper/2023/hash/a85b405ed65c6477a4fe8302b5e06ce7-Abstract-Conference.html} {Direct preference optimization: Your language model is secretly a reward model}.
\newblock In \emph{Advances in Neural Information Processing Systems 36: Annual Conference on Neural Information Processing Systems 2023, NeurIPS 2023, New Orleans, LA, USA, December 10 - 16, 2023}.

\bibitem[{Silver et~al.(2016)Silver, Huang, Maddison, Guez, Sifre, van~den Driessche, Schrittwieser, Antonoglou, Panneershelvam, Lanctot, Dieleman, Grewe, Nham, Kalchbrenner, Sutskever, Lillicrap, Leach, Kavukcuoglu, Graepel, and Hassabis}]{DBLP:journals/nature/SilverHMGSDSAPL16}
David Silver, Aja Huang, Chris~J. Maddison, Arthur Guez, Laurent Sifre, George van~den Driessche, Julian Schrittwieser, Ioannis Antonoglou, Vedavyas Panneershelvam, Marc Lanctot, Sander Dieleman, Dominik Grewe, John Nham, Nal Kalchbrenner, Ilya Sutskever, Timothy~P. Lillicrap, Madeleine Leach, Koray Kavukcuoglu, Thore Graepel, and Demis Hassabis. 2016.
\newblock \href {https://doi.org/10.1038/NATURE16961} {Mastering the game of go with deep neural networks and tree search}.
\newblock \emph{Nat.}, 529(7587):484--489.

\bibitem[{Silver et~al.(2017)Silver, Schrittwieser, Simonyan, Antonoglou, Huang, Guez, Hubert, Baker, Lai, Bolton, Chen, Lillicrap, Hui, Sifre, van~den Driessche, Graepel, and Hassabis}]{DBLP:journals/nature/SilverSSAHGHBLB17}
David Silver, Julian Schrittwieser, Karen Simonyan, Ioannis Antonoglou, Aja Huang, Arthur Guez, Thomas Hubert, Lucas Baker, Matthew Lai, Adrian Bolton, Yutian Chen, Timothy~P. Lillicrap, Fan Hui, Laurent Sifre, George van~den Driessche, Thore Graepel, and Demis Hassabis. 2017.
\newblock \href {https://doi.org/10.1038/NATURE24270} {Mastering the game of go without human knowledge}.
\newblock \emph{Nat.}, 550(7676):354--359.

\bibitem[{Tang et~al.(2024)Tang, Zhang, Wang, and Wei}]{DBLP:conf/icml/TangZWW24}
Zhengyang Tang, Xingxing Zhang, Benyou Wang, and Furu Wei. 2024.
\newblock \href {https://openreview.net/forum?id=Kjww7ZN47M} {Mathscale: Scaling instruction tuning for mathematical reasoning}.
\newblock In \emph{Forty-first International Conference on Machine Learning, {ICML} 2024, Vienna, Austria, July 21-27, 2024}. OpenReview.net.

\bibitem[{Team(2024)}]{2024internlm2wqx}
InternLM Team. 2024.
\newblock https://github.com/internlm/internlm-wqx.
\newblock \url{https://github.com/InternLM/InternLM-WQX}.

\bibitem[{Tian et~al.(2024)Tian, Peng, Song, Jin, Yu, Mi, and Yu}]{tian2024toward}
Ye~Tian, Baolin Peng, Linfeng Song, Lifeng Jin, Dian Yu, Haitao Mi, and Dong Yu. 2024.
\newblock Toward self-improvement of llms via imagination, searching, and criticizing.
\newblock \emph{arXiv preprint arXiv:2404.12253}.

\bibitem[{Tong et~al.(2024)Tong, Zhang, Wang, Wu, and He}]{DBLP:journals/corr/abs-2407-13690}
Yuxuan Tong, Xiwen Zhang, Rui Wang, Ruidong Wu, and Junxian He. 2024.
\newblock \href {https://doi.org/10.48550/ARXIV.2407.13690} {Dart-math: Difficulty-aware rejection tuning for mathematical problem-solving}.
\newblock \emph{CoRR}, abs/2407.13690.

\bibitem[{Toshniwal et~al.(2024)Toshniwal, Moshkov, Narenthiran, Gitman, Jia, and Gitman}]{DBLP:journals/corr/abs-2402-10176}
Shubham Toshniwal, Ivan Moshkov, Sean Narenthiran, Daria Gitman, Fei Jia, and Igor Gitman. 2024.
\newblock \href {https://doi.org/10.48550/ARXIV.2402.10176} {Openmathinstruct-1: {A} 1.8 million math instruction tuning dataset}.
\newblock \emph{CoRR}, abs/2402.10176.

\bibitem[{Wang et~al.(2024{\natexlab{a}})Wang, Ren, Zhou, Lu, Luo, Shi, Zhang, Song, Zhan, and Li}]{DBLP:conf/iclr/WangRZLLSZSZ024}
Ke~Wang, Houxing Ren, Aojun Zhou, Zimu Lu, Sichun Luo, Weikang Shi, Renrui Zhang, Linqi Song, Mingjie Zhan, and Hongsheng Li. 2024{\natexlab{a}}.
\newblock \href {https://openreview.net/forum?id=z8TW0ttBPp} {Mathcoder: Seamless code integration in llms for enhanced mathematical reasoning}.
\newblock In \emph{The Twelfth International Conference on Learning Representations, {ICLR} 2024, Vienna, Austria, May 7-11, 2024}. OpenReview.net.

\bibitem[{Wang et~al.(2024{\natexlab{b}})Wang, Li, Shao, Xu, Dai, Li, Chen, Wu, and Sui}]{DBLP:conf/acl/WangLSXDLCWS24}
Peiyi Wang, Lei Li, Zhihong Shao, Runxin Xu, Damai Dai, Yifei Li, Deli Chen, Yu~Wu, and Zhifang Sui. 2024{\natexlab{b}}.
\newblock \href {https://doi.org/10.18653/V1/2024.ACL-LONG.510} {Math-shepherd: Verify and reinforce llms step-by-step without human annotations}.
\newblock In \emph{Proceedings of the 62nd Annual Meeting of the Association for Computational Linguistics (Volume 1: Long Papers), {ACL} 2024, Bangkok, Thailand, August 11-16, 2024}, pages 9426--9439. Association for Computational Linguistics.

\bibitem[{Wang et~al.(2024{\natexlab{c}})Wang, Song, Tian, Yu, Peng, Mi, Huang, and Yu}]{wang2024towards}
Xiyao Wang, Linfeng Song, Ye~Tian, Dian Yu, Baolin Peng, Haitao Mi, Furong Huang, and Dong Yu. 2024{\natexlab{c}}.
\newblock Towards self-improvement of llms via mcts: Leveraging stepwise knowledge with curriculum preference learning.
\newblock \emph{arXiv preprint arXiv:2410.06508}.

\bibitem[{Wang et~al.(2023)Wang, Wei, Schuurmans, Le, Chi, Narang, Chowdhery, and Zhou}]{DBLP:conf/iclr/0002WSLCNCZ23}
Xuezhi Wang, Jason Wei, Dale Schuurmans, Quoc~V. Le, Ed~H. Chi, Sharan Narang, Aakanksha Chowdhery, and Denny Zhou. 2023.
\newblock \href {https://openreview.net/forum?id=1PL1NIMMrw} {Self-consistency improves chain of thought reasoning in language models}.
\newblock In \emph{The Eleventh International Conference on Learning Representations, {ICLR} 2023, Kigali, Rwanda, May 1-5, 2023}. OpenReview.net.

\bibitem[{Wang et~al.(2024{\natexlab{d}})Wang, Li, Wu, Luo, Hou, Yu, and Shang}]{DBLP:conf/emnlp/WangLWLH0S24}
Zihan Wang, Yunxuan Li, Yuexin Wu, Liangchen Luo, Le~Hou, Hongkun Yu, and Jingbo Shang. 2024{\natexlab{d}}.
\newblock \href {https://aclanthology.org/2024.findings-emnlp.429} {Multi-step problem solving through a verifier: An empirical analysis on model-induced process supervision}.
\newblock In \emph{Findings of the Association for Computational Linguistics: {EMNLP} 2024, Miami, Florida, USA, November 12-16, 2024}, pages 7309--7319. Association for Computational Linguistics.

\bibitem[{Wei et~al.(2022)Wei, Wang, Schuurmans, Bosma, Xia, Chi, Le, Zhou et~al.}]{wei2022chain}
Jason Wei, Xuezhi Wang, Dale Schuurmans, Maarten Bosma, Fei Xia, Ed~Chi, Quoc~V Le, Denny Zhou, et~al. 2022.
\newblock Chain-of-thought prompting elicits reasoning in large language models.
\newblock \emph{Advances in neural information processing systems}, 35:24824--24837.

\bibitem[{Wei et~al.(2023)Wei, Luan, Liu, Dong, and Wang}]{wei2023cmath}
Tianwen Wei, Jian Luan, Wei Liu, Shuang Dong, and Bin Wang. 2023.
\newblock \href {https://arxiv.org/abs/2306.16636} {Cmath: Can your language model pass chinese elementary school math test?}
\newblock \emph{Preprint}, arXiv:2306.16636.

\bibitem[{Xie et~al.(2024)Xie, Goyal, Zheng, Kan, Lillicrap, Kawaguchi, and Shieh}]{xie2024monte}
Yuxi Xie, Anirudh Goyal, Wenyue Zheng, Min-Yen Kan, Timothy~P Lillicrap, Kenji Kawaguchi, and Michael Shieh. 2024.
\newblock Monte carlo tree search boosts reasoning via iterative preference learning.
\newblock \emph{arXiv preprint arXiv:2405.00451}.

\bibitem[{Xu et~al.(2024)Xu, Sun, Zheng, Geng, Zhao, Feng, Tao, Lin, and Jiang}]{DBLP:conf/iclr/XuSZG0FTLJ24}
Can Xu, Qingfeng Sun, Kai Zheng, Xiubo Geng, Pu~Zhao, Jiazhan Feng, Chongyang Tao, Qingwei Lin, and Daxin Jiang. 2024.
\newblock \href {https://openreview.net/forum?id=CfXh93NDgH} {Wizardlm: Empowering large pre-trained language models to follow complex instructions}.
\newblock In \emph{The Twelfth International Conference on Learning Representations, {ICLR} 2024, Vienna, Austria, May 7-11, 2024}. OpenReview.net.

\bibitem[{Yang et~al.(2024{\natexlab{a}})Yang, Yang, Zhang, Hui, Zheng, Yu, Li, Liu, Huang, Wei, Lin, Yang, Tu, Zhang, Yang, Yang, Zhou, Lin, Dang, Lu, Bao, Yang, Yu, Li, Xue, Zhang, Zhu, Men, Lin, Li, Xia, Ren, Ren, Fan, Su, Zhang, Wan, Liu, Cui, Zhang, and Qiu}]{DBLP:journals/corr/abs-2412-15115}
An~Yang, Baosong Yang, Beichen Zhang, Binyuan Hui, Bo~Zheng, Bowen Yu, Chengyuan Li, Dayiheng Liu, Fei Huang, Haoran Wei, Huan Lin, Jian Yang, Jianhong Tu, Jianwei Zhang, Jianxin Yang, Jiaxi Yang, Jingren Zhou, Junyang Lin, Kai Dang, Keming Lu, Keqin Bao, Kexin Yang, Le~Yu, Mei Li, Mingfeng Xue, Pei Zhang, Qin Zhu, Rui Men, Runji Lin, Tianhao Li, Tingyu Xia, Xingzhang Ren, Xuancheng Ren, Yang Fan, Yang Su, Yichang Zhang, Yu~Wan, Yuqiong Liu, Zeyu Cui, Zhenru Zhang, and Zihan Qiu. 2024{\natexlab{a}}.
\newblock \href {https://doi.org/10.48550/ARXIV.2412.15115} {Qwen2.5 technical report}.
\newblock \emph{CoRR}, abs/2412.15115.

\bibitem[{Yang et~al.(2024{\natexlab{b}})Yang, Zhang, Hui, Gao, Yu, Li, Liu, Tu, Zhou, Lin, Lu, Xue, Lin, Liu, Ren, and Zhang}]{DBLP:journals/corr/abs-2409-12122}
An~Yang, Beichen Zhang, Binyuan Hui, Bofei Gao, Bowen Yu, Chengpeng Li, Dayiheng Liu, Jianhong Tu, Jingren Zhou, Junyang Lin, Keming Lu, Mingfeng Xue, Runji Lin, Tianyu Liu, Xingzhang Ren, and Zhenru Zhang. 2024{\natexlab{b}}.
\newblock \href {https://doi.org/10.48550/ARXIV.2409.12122} {Qwen2.5-math technical report: Toward mathematical expert model via self-improvement}.
\newblock \emph{CoRR}, abs/2409.12122.

\bibitem[{Yao et~al.(2024)Yao, Yu, Zhao, Shafran, Griffiths, Cao, and Narasimhan}]{yao2024tree}
Shunyu Yao, Dian Yu, Jeffrey Zhao, Izhak Shafran, Tom Griffiths, Yuan Cao, and Karthik Narasimhan. 2024.
\newblock Tree of thoughts: Deliberate problem solving with large language models.
\newblock \emph{Advances in Neural Information Processing Systems}, 36.

\bibitem[{Ying et~al.(2024)Ying, Zhang, Li, Zhou, Shao, Fei, Ma, Hong, Liu, Wang, Wang, Wu, Li, Zhou, Liu, Zhang, Zhang, Yan, Qiu, Wang, Chen, and Lin}]{ying2024internlmmathopenmathlarge}
Huaiyuan Ying, Shuo Zhang, Linyang Li, Zhejian Zhou, Yunfan Shao, Zhaoye Fei, Yichuan Ma, Jiawei Hong, Kuikun Liu, Ziyi Wang, Yudong Wang, Zijian Wu, Shuaibin Li, Fengzhe Zhou, Hongwei Liu, Songyang Zhang, Wenwei Zhang, Hang Yan, Xipeng Qiu, Jiayu Wang, Kai Chen, and Dahua Lin. 2024.
\newblock \href {https://arxiv.org/abs/2402.06332} {Internlm-math: Open math large language models toward verifiable reasoning}.
\newblock \emph{Preprint}, arXiv:2402.06332.

\bibitem[{Yu et~al.(2023)Yu, Jiang, Shi, Yu, Liu, Zhang, Kwok, Li, Weller, and Liu}]{yu2023metamath}
Longhui Yu, Weisen Jiang, Han Shi, Jincheng Yu, Zhengying Liu, Yu~Zhang, James~T Kwok, Zhenguo Li, Adrian Weller, and Weiyang Liu. 2023.
\newblock Metamath: Bootstrap your own mathematical questions for large language models.
\newblock \emph{arXiv preprint arXiv:2309.12284}.

\bibitem[{Yu et~al.(2024)Yu, Jiang, Shi, Yu, Liu, Zhang, Kwok, Li, Weller, and Liu}]{DBLP:conf/iclr/YuJSYLZKLWL24}
Longhui Yu, Weisen Jiang, Han Shi, Jincheng Yu, Zhengying Liu, Yu~Zhang, James~T. Kwok, Zhenguo Li, Adrian Weller, and Weiyang Liu. 2024.
\newblock \href {https://openreview.net/forum?id=N8N0hgNDRt} {Metamath: Bootstrap your own mathematical questions for large language models}.
\newblock In \emph{The Twelfth International Conference on Learning Representations, {ICLR} 2024, Vienna, Austria, May 7-11, 2024}. OpenReview.net.

\bibitem[{Yuan et~al.(2023)Yuan, Yuan, Li, Dong, Lu, Tan, Zhou, and Zhou}]{yuan2023scalingrelationshiplearningmathematical}
Zheng Yuan, Hongyi Yuan, Chengpeng Li, Guanting Dong, Keming Lu, Chuanqi Tan, Chang Zhou, and Jingren Zhou. 2023.
\newblock \href {https://arxiv.org/abs/2308.01825} {Scaling relationship on learning mathematical reasoning with large language models}.
\newblock \emph{Preprint}, arXiv:2308.01825.

\bibitem[{Yue et~al.(2023)Yue, Qu, Zhang, Fu, Huang, Sun, Su, and Chen}]{DBLP:journals/corr/abs-2309-05653}
Xiang Yue, Xingwei Qu, Ge~Zhang, Yao Fu, Wenhao Huang, Huan Sun, Yu~Su, and Wenhu Chen. 2023.
\newblock \href {https://doi.org/10.48550/ARXIV.2309.05653} {Mammoth: Building math generalist models through hybrid instruction tuning}.
\newblock \emph{CoRR}, abs/2309.05653.

\bibitem[{Zhang et~al.(2024{\natexlab{a}})Zhang, Zhoubian, Hu, Yue, Dong, and Tang}]{zhang2024rest}
Dan Zhang, Sining Zhoubian, Ziniu Hu, Yisong Yue, Yuxiao Dong, and Jie Tang. 2024{\natexlab{a}}.
\newblock Rest-mcts*: Llm self-training via process reward guided tree search.
\newblock \emph{arXiv preprint arXiv:2406.03816}.

\bibitem[{Zhang et~al.(2024{\natexlab{b}})Zhang, Huang, Zhou, Li, and Ouyang}]{zhang2024accessing}
Di~Zhang, Xiaoshui Huang, Dongzhan Zhou, Yuqiang Li, and Wanli Ouyang. 2024{\natexlab{b}}.
\newblock Accessing gpt-4 level mathematical olympiad solutions via monte carlo tree self-refine with llama-3 8b.
\newblock \emph{arXiv preprint arXiv:2406.07394}.

\bibitem[{Zheng et~al.(2023)Zheng, Yin, Xie, Huang, Sun, Yu, Cao, Kozyrakis, Stoica, Gonzalez, Barrett, and Sheng}]{DBLP:journals/corr/abs-2312-07104}
Lianmin Zheng, Liangsheng Yin, Zhiqiang Xie, Jeff Huang, Chuyue Sun, Cody~Hao Yu, Shiyi Cao, Christos Kozyrakis, Ion Stoica, Joseph~E. Gonzalez, Clark~W. Barrett, and Ying Sheng. 2023.
\newblock \href {https://doi.org/10.48550/ARXIV.2312.07104} {Efficiently programming large language models using sglang}.
\newblock \emph{CoRR}, abs/2312.07104.

\end{thebibliography}
\newpage
\appendix

\section{Node Defination}
\label{apd:node}
In our application, each reasoning step is represented as a node. The reasoning trajectories are segmented into individual steps according to strings such as "Step 1", "Step 2", and so on, which are automatically generated by the model in its output. Each step may consist of a single sentence or span multiple sentences. Through the use of few-shot examples, the model learns to follow this structured format when generating multi-step reasoning paths, thereby enabling node-level partitioning. Below is one example of how each node is defined.

% 正文内容

\begin{stepbox}[title=Root(Query)]
Real numbers $x$ and $y$ with $x,y>1$ satisfy $\log_x(y^x)=\log_y(x^{4y})=10.$ What is the value of $xy$?

Please solve this task step by step and use the format like "Step 1:", "Step 2:" to structure your Chain of Thought (CoT) steps. Now, generate your solution starting from Step 1:
\end{stepbox}

\begin{stepbox}
Step 1: Start by analyzing the given equations. We have $\log_x(y^x) = 10$ and $\log_y(x^{4y}) = 10$.
\end{stepbox}

\begin{stepbox}
Step 2: Use the property of logarithms that $\log_b(a^c) = c\log_b(a)$ to simplify the equations. For the first equation, we have $\log_x(y^x) = x\log_x(y) = 10$. For the second equation, we have $\log_y(x^{4y}) = 4y\log_y(x) = 10$.
\end{stepbox}

\begin{stepbox}
Step 3: From Step 2, we can express $\log_x(y)$ and $\log_y(x)$ in terms of $x$ and $y$:
\[
\log_x(y) = \frac{10}{x}
\]
\[
log_y(x) = \frac{10}{4y} = \frac{5}{2y}
\]
\end{stepbox}

\begin{stepbox}
Step 4: Use the change of base formula $\log_a(b) = \frac{1}{\log_b(a)}$ to relate $\log_x(y)$ and $\log_y(x)$:
\[
\log_x(y) = \frac{1}{\log_y(x)}
\]
Substituting the values from Step 3, we get $\frac{10}{x} = \frac{1}{\frac{5}{2y}}$.

\end{stepbox}

\begin{stepbox}
Step 5: Simplify the equation from Step 4:
\[
\frac{10}{x} = \frac{2y}{5}
\]
Cross multiply to get $50 = 2xy$

Divide both sides by 2 to solve for $xy$: $xy = 25$.
\end{stepbox}

\begin{stepbox}
Step 6: Conclude that the value of $xy$ is 25.
Thus, the answer is $\boxed{25}$.
\end{stepbox}
\section{FastMCTS Algorithms}
\label{apd:fastmcts}

\begin{algorithm*}
\caption{FastMCTS}
\label{algo:fastmcts}
\KwIn{Input query  $q$, ground truth $g$, few shot set $\mathcal{F}$, policy model $\pi_\theta$, verifier model $V_\phi$, initial degree $d_i$, expand degree $d_e$, iterations $N$, difficulty thresholds $l_{high}, l_{low}$, UCT constant $c$ }
\KwOut{The search tree $T$ of input query $q$}

\textbf{Initialize:} search tree $T$ with root $\gets$ $q$

% \COMMENTLIGHTGRAY{test for gray comment}

%\texttt{init\_process\_group(world\_size=2)}\\

% $prefix\_unverified\gets \texttt{None}$\\

\While{iter < N}{
\COMMENTLLAMA{Recursively select node with Adaptive Stay Policy}

% select from root
current\_node $\gets$ root

selected\_node $\gets$ None
    
\While{\textnormal{selected\_node is None}}{

candidate\_children $\gets$ current\_node.children

\If{\textnormal{number of candidate\_children} $<=$ 1 \texttt{or} \COMMENTLIGHTGRAY{\textnormal{Adaptive Stay Policy}}
\quad \textnormal{all candidate\_children are leaf nodes} \texttt{or} \\
\quad current\_node.visit\_count > 1 \texttt{and} current\_node.score $\in (0, l_{\text{low}}] \cup [l_{\text{high}}, 1)$ 
}{
selected\_node $\gets$ current\_node

break
}

\eIf{current\_node.visit\_count > 1}{$c_{current}$ $\gets$ $c \cdot current\_node.score$ \COMMENTLIGHTGRAY{Dynamic Exploration}} 
{$c_{current}$ $\gets$ $c$}

candidate\_node $\gets$ $\arg\max_{node \in candidate\_children} UCT(node,c_{current})$ 

\If{candidate\_node.visit\_count > 1 \texttt{and} candidate\_node.score <= $l_{low}$}{
selected\_node $\gets$ candidate\_node
}

current\_node $\gets$ candidate\_node

}

\COMMENTLLAMA{Expansion and Simulation}
% get $prob_{draft}, $ via inter-process communication\\
Get current state $s_t$ from root to selected\_node: \( s_t = (a_t, a_{t-1}, \dots, a_1,q) \) \\

\eIf{\textnormal{candidate\_node is root} }
{   
    Sample $d_i$ partial trajectories  $\left\{\boldsymbol{\tau}^{(i)}\right\}_{i=1}^{d_i} \sim \pi_{\theta}(\boldsymbol{\tau}\mid s_t, f^{(i)}),\quad f^{(i)} \subseteq \mathcal{F} $ \\
    \COMMENTLIGHTGRAY{Sample with random fewshot}
}
{
    Sample $d_e$ partial trajectories  $\left\{\boldsymbol{\tau}^{(i)}\right\}_{i=1}^{d_e} \sim \pi_{\theta}(\boldsymbol{\tau}\mid s_t, f^{(i)}),\quad f^{(i)} \subseteq \mathcal{F} $
}
\textbf{Split} $\left\{\boldsymbol{\tau}^{(i)}\right\}$ to multi steps 
$\left\{(a_{t+1}^{(i)}, a_{t+2}^{(i)}, \dots, a_{end}^{(i)}) \right\}$ and construct them as new branches of tree nodes $\left\{(node_{t+1}^{(i)}, node_{t+2}^{(i)}, \dots, node_{end}^{(i)}) \right\}$\\
\textbf{Append} these new branches to selected\_node \COMMENTLIGHTGRAY{Reserve all simulation results}

\COMMENTLLAMA{Backup}

Use verifier model $V_\phi$ to judge $\left\{\boldsymbol{\tau}^{(i)}\right\}$  \COMMENTLIGHTGRAY{Use LLM as verifer}

\textbf{Backup} score from newly expanded tree nodes using Monte Carlo Evaluation \\

}

\end{algorithm*}

% 这里可以补充一下，根据那些参数能采样到多少条数据？

The full FastMCTS algorithm is outlined in Algorithm \ref{algo:fastmcts}.

\begin{table*}[htb]
\footnotesize
\centering

\begin{tabular}{@{}>{\raggedright}p{3.2cm}ccccccccc@{}}
\toprule
Model Name           & AIME24 & MATH & GSM8K & AMC23 & \begin{tabular}[c]{@{}c@{}}Olympiad\\ Bench\end{tabular} & OmniMath & \begin{tabular}[c]{@{}c@{}}SAT\\ Math\end{tabular} & \begin{tabular}[c]{@{}c@{}}Gaokao\\ Math\end{tabular} & Avg. \\ 
\midrule
Llama3.2\_3b\_base\_RS         &   0.0  & 31.0 &  55.5 & 15.0  &      10.7      &   10.7   &   50.2   &    30.0     & 25.4 \\
Llama3.2\_3b\_base\_FastMCTS    &   3.0  & 35.2 &  53.4 & 15.0  &      12.6      &   11.4   &   50.0   &    32.9     & \textbf{26.7} \\
Qwen2.5\_3b\_base\_RS          &   6.7  & 62.2 &  83.6 & 35.0  &      27.0      &   20.8   &   70.6   &    56.1     & 45.3 \\
Qwen2.5\_3b\_base\_FastMCTS     &  10.0  & 62.2 &  83.3 & 45.0  &      29.5      &   21.5   &   71.5   &    56.1     & \textbf{47.4} \\
Qwen2.5\_7b\_base\_RS          &   6.7  & 72.0 &  89.1 & 52.5  &      27.6      &   38.3   &   70.6   &    62.6     & 52.4 \\ 
Qwen2.5\_7b\_base\_FastMCTS     &  13.3  & 73.0 &  88.9 & 57.5  &      28.1      &   39.8   &   74.5   &    63.6     & \textbf{54.8} \\
\bottomrule
\end{tabular}
\caption{The results of different model performance when trained with data generated by Rejection Sampling and FastMCTS.}
\label{tab:different}
\end{table*}

\section{Details of Training Data Selection}
\label{apd:data}
In our preliminary experiments to evaluate the efficiency of rejection sampling, we employed Qwen2.5-32B-Instruct to sample responses across our full Chinese and English dataset, with 5 samples generated per question. This allowed us to calculate a "pass rate" (percentage of correct solutions) for each question, which we used to stratify problem complexity. Questions with a 100\% pass rate were deemed excessively simple and excluded from the dataset.

Additionally, we applied heuristic filtering strategies to remove ambiguous or low-quality problems:

\begin{itemize}
    \item Rule-based exclusion: Problems containing keywords such as "proof", "prove", "show that", "find all" , or url/image extensions (e.g., "http", ".png", ".jpg", "www", ".svg", ".bmp") were automatically filtered out.
    \item Format checks : Questions with formatting errors (e.g., broken LaTeX, incomplete sentences) were discarded.
    \item Deduplication : We removed duplicate entries via hash-based matching and ensured no overlap with the test set.
\end{itemize}

\section{Details of Model Evaluation}
\label{apd:judge}
As we have mentioned in Section \ref{sec:robust}, we propose to employ an LLM to verify the correctness of each reasoning path, aiming to identify logical
errors and exclude trajectories that are guessed answers. For prompt design of LLM judge, an example prompt template is demonstrated in Figure \ref{fig:prompt_example}.

\begin{figure*}[t]
\begin{Verbatim}[frame=single, fontsize=\footnotesize]
##Question##
{question}

##Student's Answer##
{model_output}

The standard answer for this question is as follows:
##Standard Answer##
{answer}

Now, based on the standard answer, determine whether the student's answer is correct. 
(Please note that the same mathematical expression may have different formats or equivalent forms).
You only need to focus on:
1. Whether the student's answer matches the result of the standard answer.
2. Whether the student's answer seems to be guessed or is a vague answer. If the student's answer 
is correct (if there are multiple questions, all sub-questions must be answered correctly), 
please reply directly with:
**Correct Answer**
If the student's answer is incorrect, please reply directly with:
**Incorrect Answer**
\end{Verbatim}
\caption{Example of the Prompt Template Used for Model Evaluation}
\label{fig:prompt_example}
\end{figure*}

Meanwhile, to reduce computational costs, we limited the maximum output length to 32 tokens (as only final answers are required). To ensure accuracy, we employed a majority voting strategy: the judge model verifies each answer N=3 times , and only consistent results across all trials are accepted. If inconsistencies arise, the verification is repeated until consensus is reached. This approach minimizes errors and outperforms rule-based matching in identifying nuanced correct answers.

The rigorous validation was critical because the synthesized data is used not only for supervised fine-tuning but also for Branch-DPO, where precise step-level evaluation (distinguishing true positives/negatives) is essential for Preference Optimization. All described details have been fully implemented in our code.

\section{Sampling Settings}
\label{apd:sampling settings}

For all our sampling settings, we use SGLang~\cite{DBLP:journals/corr/abs-2312-07104} as our inference engine and employ sampling generation with a temperature setting of 1 to ensure diversity. In FastMCTS, the constant $c$ in the UCT score is set to its default value of 1.414. Additionally, we utilize Qwen2.5-72B-Instruct as a LLM judger to verify the solutions. 

We use an asynchronous approach in our implementation, allowing different branches of the search tree to be processed concurrently. Although FastMCTS requires multiple iterations to construct a search tree for each problem, this parallel processing allows us to perform inference on a large number of inputs simultaneously, thereby ensuring high efficiency.

In section \ref{sec:efficiency}, to scale up the sampling computation, for FastMCTS, we incrementally increased the number of iterations from 4 to 20, and the expansion degree (i.e., the number of nodes expanded after the selection phase) is varied from 1 to 2. For Rejection Sampling, we expanded the number of generated trajectories per query from 3 to 32.

In section \ref{sec:performance}, to obtain comparable sampling computation, for each query in the original dataset, we sampled multiple times (30 for English data and 24 for Chinese data) using rejection sampling. For FastMCTS, it starts with an initial degree of 3 at the root, expands by adding 2 branches in each expansion phase, and performs 16 iterations of tree search.

\section{Training Data Construction}
\label{apd:data construct}
\paragraph{Supervised Fine-tuning}

After the sampling process, each problem is sampled with varying numbers of solution candidates. To investigate the impact of both training data size and the number of reasoning trajectories per problem, we impose constraints on the maximum number of solutions utilized per problem during the training process. This approach also helps maintain a balance between different problems.

For Rejection Sampling, we selecte correct trajectories for each problem randomly. For FastMCTS, our strategy involves prioritizing the selection of correct trajectories from various branches of the search tree. By doing so, we aim to maximize the diversity of the training data.
% The sizes of the supervised fine-tuning datasets obtained by both sampling methods and their corresponding training results are shown in Table \ref{tab:train res}.

\paragraph{Branch-DPO}

In addition to improving the efficiency of sampling correct reasoning paths, FastMCTS also provides step-level supervision information. Unlike rejection sampling, which generates multiple completely independent trajectories for each problem, FastMCTS constructs a search tree for each problem, where each node stores a score computed through Monte Carlo evaluation. This allows for step-level or branch-level preference optimization based on the scores of tree nodes.

% 这里是否需要放DPO公式？
% 附录是不是要放一下构造dpo数据的算法？因为如何构造dpo数据不算太重要？
Direct Preference Optimization (DPO)~\cite{DBLP:conf/nips/RafailovSMMEF23} has been widely adopted for model optimization due to its efficiency in utilizing pairwise preference data. It has also been applied to step-level preference optimization, as most undesirable trajectories do not initially contain errors~\cite{lai2024stepdpostepwisepreferenceoptimization,xie2024monte,DBLP:journals/corr/abs-2405-03553,wang2024towards}. 

We propose a simple algorithm to construct preference data from the tree structures generated by FastMCTS. Our approach is based on the following assumptions: 

1. For a multi-step reasoning trajectory, if the final result is correct and clear, all intermediate steps are considered correct.

2. If the final result is incorrect, the intermediate steps are not necessarily incorrect. 

However, if a step has been simulated multiple times and its Monte Carlo-estimated score remains zero, it can be considered a \textbf{"low-quality node."} Based on this, we construct step-level or branch-level preference data. For any node in the tree, we examine its child nodes. If a child node is identified as low-quality, we construct step-level preference data between this node and a high-quality node that has led to correct results. If the child branches contain both correct and incorrect results but have only been simulated once, we cannot definitively assess the quality of individual steps and instead construct branch-level preference data.

 In our experiments, for each search tree associated with one problem, we construct up to 5 step-level or branch-level preference pairs, resulting in an additional 152K(on CN High School Math Hard) and 215K(on En Math Hard) preference data points for DPO training. This approach further leverages the tree-structured data generated by FastMCTS.

\section{Training Setups}
\label{apd:training details}
We use Qwen2.5-7B as our base model and perform training on datasets generated by both FastMCTS and rejection sampling. For supervised fine-tuning, the maximum sequence length is set to 4096 tokens, and the global batch size is set to 32. We employ the Adam optimizer with a learning rate of 1e-5 and a linear warmup schedule with a warmup step ratio of 0.1. For all synthetic datasets, we train the model for 3 epochs and select the best checkpoint based on validation performance.

After supervised fine-tuning, we further refine the best checkpoint trained on FastMCTS-generated data using Branch-DPO for 3 epochs. The global batch size for Branch-DPO is set to 16, and the learning rate is set to 1e-6. The hyperparameter \(\beta\) is set to 0.4. We use the AdamW optimizer with a cosine learning rate scheduler and a warmup ratio of 0.1.

\section{Train with Different Models}
\label{apd:different models}

We also evaluated our methods on LLMs of different series and sizes. Using the same experimental setup as in Section \ref{sec:performance}, we evaluated the En Math Hard dataset with less than 5 reasoning trajectories per problem. We compared the fine-tuning results(best checkpoint in 3 epochs) of synthetic data generated by FastMCTS and Rejection Sampling using two additional base models: Llama-3.2-3b-base and Qwen2.5-3b-base. The results are shown in Table \ref{tab:different}:

These results demonstrate that even with different base models, under the same synthetic data cost, fine-tuning with data generated by FastMCTS consistently outperforms Rejection Sampling.

\section{Performance against Recent Works}

Our work proposes an algorithm designed to improve the efficiency and quality of synthetic data sampling for reasoning paths. Regarding the original data we selected, we primarily leverage open-source dataset and problems collected from website, this may not be the optimal instruction data for training SOTA models. However, we also compare our model performance compared to recent works with comparable model size, which is described in Table \ref{tab:sota}.

These results suggest that our model achieves competitive performance. We recognize that incorporating higher-quality dataset curation could further improve outcomes, which is a direction we plan to explore in future work.

\begin{table}[htbp]
\centering
\scriptsize

\begin{tabular}{lcccc}
\hline
Model Name & AIME24 & MATH & GSM8K & AMC23 \\
\hline
GPT-4o & 9.3 & 76.6 & 92.9 & 47.5 \\
NuminaMath-CoT-7B & 0 & 55.8 & 76.3 & 27.5 \\
NuminaMath-TIR-7B & 16.7 & 68.1 & 84.6 & 50.0 \\
OpenMath2-Llama3.1-8B & 10.0 & 67.8 & 91.7 & 40.0 \\
rStarMath Policy 7B & 26.7 & 78.4 & 89.7 & 47.5 \\
FastMCTS 7B  & 20.0 & 75.4 & 89.9 & 57.5 \\
\hline
\end{tabular}
\caption{Performance comparison with recent works on various benchmarks.}
\label{tab:sota}
\end{table}

\end{document}